%% file: main.tex
\documentclass[12pt]{article}

\usepackage{rotating}
\usepackage{multirow}
\usepackage{amsmath}
\usepackage{amssymb}
\usepackage{amsthm}
\usepackage{thm-restate} 
\usepackage{graphicx}
\usepackage{url}
\usepackage{makecell} 
\usepackage{blkarray}
\usepackage{subfiles}
\usepackage[authoryear]{natbib}
\usepackage{placeins}
\usepackage[hidelinks]{hyperref}
\usepackage{bbm}
\usepackage[export]{adjustbox}
\usepackage{arydshln}
\usepackage{appendix}
\usepackage{caption}

\usepackage[dvipsnames]{xcolor} 

\usepackage{fancyhdr}
\usepackage{thmtools} 

\declaretheoremstyle[
  bodyfont=\normalfont,
  spaceabove=1em plus 0.75em minus 0.25em,
  spacebelow=2em plus 0.75em minus 0.25em,
  qed={$\triangle$},
]{exmpstyle2}
\newtheorem{theorem}{Theorem}
\newtheorem{corollary}{Corollary}

\newtheorem{remark}{Remark}
\theoremstyle{definition}\newtheorem{definition}{Definition}
\newtheorem{process}[definition]{Process}
\declaretheorem[
  style=exmpstyle2,
  title=Example,
  refname={example,examples},
  Refname={Example,Examples}
]{example}

\graphicspath{ {./figures/} }

\newcommand{\myvec}[1]{\underline{\boldsymbol{{#1}}}}
\newcommand{\mymatrix}[1]{\boldsymbol{#1}}

\newcommand{\subheader}[1]{\vspace{0.15cm}\noindent\textit{#1}.}

\renewcommand{\quote}[1]{`#1'}

\newcommand{\oo}[1]{[#1]} 

\newcommand{\modelParams}{\text{Model Parameters}}
\newcommand{\nbaseclass}{\mathcal{B}}
\newcommand{\formattable}[1]
{{\spacingset{1}\centerline{\adjustbox{max width=\textwidth,keepaspectratio}{#1}}}}
\newcommand{\myrotate}[3]{\makecell[#1]{\begin{rotate}{#2} #3 \end{rotate}}}
\DeclareRobustCommand{\stirling}{\genfrac\{\}{0pt}{}}
\newcommand{\khatrirao}{*} 
\newcommand\numberthis{\addtocounter{equation}{1}\tag{\theequation}}

\def\spacingset#1{\renewcommand{\baselinestretch}{#1}\small\normalsize\everydisplay{\def\arraystretch{0.5}}} 

\title{Equivalence Set Restricted Latent Class Models (ESRLCM)}
\author{Jesse Bowers; Steve Culpepper}

\begin{document}

\pagenumbering{roman}

\maketitle

\section{Abstract}

Latent Class Models (LCMs) are used to cluster multivariate categorical data, commonly used to interpret survey responses. We propose a novel Bayesian model called the Equivalence Set Restricted Latent Class Model (ESRLCM). This model identifies clusters who have common item response probabilities, and does so more generically than traditional restricted latent attribute models. We verify the identifiability of ESRLCMs, and demonstrate the effectiveness in both simulations and real-world applications.

\begingroup
\setcounter{tocdepth}{1}
\renewcommand*{\addvspace}[1]{}
\tableofcontents
\endgroup
\newpage
\pagenumbering{arabic}

\spacingset{1}

\subfile{PaperSections/0100_introduction.tex}

\subfile{PaperSections/0200_restrictions.tex}

\subfile{PaperSections/0250_identifiability.tex}

\subfile{PaperSections/0300_dependentBeta.tex}

\subfile{PaperSections/0500_posteriors.tex}

\subfile{PaperSections/0600_simulations.tex}

\subfile{PaperSections/0700_applications.tex}

\subfile{PaperSections/0720_applicationMatrices.tex}

\subfile{PaperSections/1000_conclusion.tex}

\bibliographystyle{plainnat} 

\input{bibliography.bbl}
\newpage
\appendix

\subfile{Appendix/0200_restrictions.tex}

\subfile{Appendix/0300_dependentBeta.tex}

\subfile{Appendix/0500_posteriors.tex}

\subfile{Appendix/0600_simulations.tex}

\subfile{Appendix/0710_applicationFractionSubtraction.tex}

\section{End} 


\end{document}

%% file: PaperSections/0100_introduction.tex
\FloatBarrier

\section{Introduction}

Researchers continue to use latent class models (LCMs) to provide a parsimonious representation of multivariate categorical data. For each observation $i$ we observe $J$ items forming a multivariate nominal response: $\myvec{X}_{i}\in \mathbb{Z}_{m_{1}}\times \cdots \times \mathbb{Z}_{m_J}$ with $\mathbb{Z}_{k}:=\{1,\cdots,k\}$. The distribution of $\myvec{X}_{i}$ is given by a mixture:
\begin{equation}
P(\myvec{x}_{i}) 
= \sum_{c\in\mathbb{Z}_{C}} P(c_{i}=c) P(\myvec{x}_{i}|c_{i}=c)
= \sum_{c\in\mathbb{Z}_{C}} P(c_{i}=c) \prod_{j \in \mathbb{Z}_{J}} P(x_{ij}|c_{i}=c)
\end{equation}
where $p(\myvec{x}|c_{i})$ is the conditional distribution of $\myvec{x}_{i}$ given the hidden class $c_{i}\in \mathbb{Z}_{C}$ and we consider the situation where the elements of $\myvec x$ are conditionally independent given class membership. We let $\myvec{\pi} := [P(c_{i}=1),\cdots,P(c_{i}=C)]^{\top}$ denote the prior probability of residing within each class, with the total number of classes $C$ fixed.

In practice a pair of classes may share qualities in common. In particular, two classes might share common response probabilities $P(x_{ij}|c_{i}=c)$ for a subset of items. For example, the equivalence of response probabilities for classes in educational applications provide researchers with useful fine-grained diagnostic information \cite[e.g., see][]{chen2015statistical,liu2013theory,xu2017identifiability}. Traditional LCMs assume each class is entirely distinct and have difficulty recovering this structure. There are at least two existing methods to recover common response probabilities. One strategy is to employ regularized latent class models, for instance applying lasso to class response probabilities \citep{Chen2016,Robitzsch2020} or tree-based regularization \citep{Li2023}. Another strategy employs restricted latent attribute models (RLAMs), where classes are encoded as binary attributes to enforce a simpler structure in class response probabilities \citep{Chen2020,Gu2024,xu2018identifying}.

We propose a Bayesian variant of the frequentist regularized latent class models. We call this new Bayesian model the Equivalence Set Restricted Latent Class Model (ESRLCM). Like other regularized LCMs, ESRLCMs provide flexible structure to capture common class response probabilities. These models can impose restrictions unavailable to restricted latent attribute models. This paper offers the following contributions.
\begin{enumerate}
\item Compared to regularized LCMs by \cite{Chen2016,Robitzsch2020}, ESRLCMs are Bayesian instead of frequentist. This allows for convenient measurements of credible intervals, including measuring the posterior probability that two classes have a common response probability. Compared with Bayesian ESRLCM by \cite{Li2023}, our model encourages identical class response probabilities rather than similar probabilities; this allows easier interpretation.
\item We provide new identifiability theory which admits these more flexible structures. The new identifiability conditions apply to other regularized LCMs as well as certain restricted latent attribute models.
\item We introduce a new hierarchical mixture prior which encourages well separated \citep{Liu2010} latent classes. Specifically, the prior encourages similar response probabilities to be equal and distinct probabilities to be divergent.
\item We compare our model with several existing competitors and demonstrate its effectiveness in applications involving educational and psychological data as well as simulations.
\end{enumerate}

Our paper is organized as follows. Section~3 provides background information on restricted latent attribute models (RLAMs) and Section~4 introduces equivalence set restrictions and contrast them with RLAMs. In Section~5, we establish identifiability of ESRLCMs. We introduce a novel prior which promotes separation in class response probabilities in Section~6 and we present details concerning the the conditional distributions and a Monte Carlo Markov chain algorithm in Section~7. In Section~8 and Section~9, we show the effectiveness of ESRLCMs in simulation studies and real world applications respectively.

\section{Restricted Latent Attribute Models}

In this section we describe two types of restricted latent attribute models (RLAMs). RLAMs express a person's latent class as a binary vector $\myvec{\alpha}_{i} \in \{0,1\}^{M}$. Each component of this vector is called a latent attribute. The foundational assumption is that each item is influenced by some latent attributes, but not impacted by others.

The first type of RLAMs use $\mymatrix{Q}$-matrix restrictions \cite[]{de2011generalized,henson2009defining,von2008general}. Under this model, matrix $\mymatrix{Q} \in \{0,1\}^{J\times K}$ indicates the relationships between observed items and latent attributes.  For instance, in the case where $X_{ij}\in \{0,1\}$ the item response probability can be written as a generalized linear model,
\begin{align}
\text{logit } P(X_{ij}|\myvec{\alpha}_{i},\mymatrix{\delta},\myvec{\beta}_{j})
&= \beta_{j0} + \sum_{k=1}^{K} \beta_{j,k} Q_{jk}\alpha_{k} + \sum_{t,t'=1}^{K} \beta_{j,t+Kt'} Q_{jt}Q_{jt'} \alpha_{t}\alpha_{t'}\notag\\
&\quad + \cdots + \beta_{j,2^{K}-1}\prod_{k} Q_{jk}\alpha_{k}\label{eq:gdm}
\end{align}
where all the main- and interaction-effect terms involving the attributes are included as predictors and the $2^K$-vector $\myvec{\beta}_j$ are the item parameters. Equation \eqref{eq:gdm} shows that the elements of $\mymatrix{Q}$ serve the role of variable selection. Specifically, if $Q_{j k}=1$, then item $j$ is dependent on attribute $k$. Conversely if $Q_{jk}=0$ then item $j$ is independent of attribute $k$.

The second type of RLAMs use $\mymatrix{\delta}$-matrix restrictions \cite[]{chen2015statistical,Chen2020}, which can be viewed as an extension of $\mymatrix{Q}$-matrix restrictions. Here matrix $\mymatrix{\delta} \in \{0,1\}^{J \times 2^{K}}$ indicates the relationship between items and latent attributes such that $\delta_{jp}$ indicates whether $\beta_{jp}$ is an active non-zero coefficient. 
In this case, the class response probabilities are given by:
\begin{align}
\text{logit } P(X_{ij}|\myvec{\alpha}_{i},\beta)
= \beta_{j0} + \sum_{k=1}^{K} \beta_{j,k} \alpha_{k} + \sum_{t,t'=1}^{K} \beta_{j,t+Kt'} \alpha_{t}\alpha_{t'} + \cdots + \beta_{j,2^{K}-1}\prod_{k} \alpha_{k}\label{eq:deltagdm}
\end{align}
with $\beta_{j t}$ restricted to zero when $\delta_{j t}=0$. In this way, the class response probabilities depend not just on which attributes are active, but also may depend on a combination of attributes.

%% file: PaperSections/0200_restrictions.tex
\FloatBarrier
\section{Equivalence Set Restrictions}\label{section:eqsets}

Although ESRLCMs are most similar to regularized LCMs, they use a different mechanism to coerce common class response probabilities. Regularized LCMs typically use a penalty (e.g. lasso), and restricted latent attribute models use $\mymatrix{\delta}$ or $\mymatrix{Q}$ restrictions. ESRLCMs use a new type of restriction to build common response probabilities: equivalence set restrictions. Equivalence set restrictions provide the same amount of flexibility as other regularized LCMs when identifying common response probabilities. In this section we detail equivalence set restrictions and contrast them with RLAM type restrictions. 

\begin{definition}[Equivalence set restrictions] Let $\mymatrix{B} \in \mathbb{Z}_{C}^{C \times J}$ be our base class matrix. If $B_{c_{1} j}=B_{c_{2} j}$ this implies that:
\begin{align*}
P(x_{ij}|c_{i}=c_{1},\modelParams)&=P(x_{ij}|c_{i}=c_{2},\modelParams).
\end{align*}
\end{definition}

\begin{definition}
The `equivalence set' for variable $j$ $E_{j b} := \{c : B_{c j}=b\} \subseteq \mathbb{Z}_{c}$ specifies a group of classes containing common response probabilities.
\end{definition}

The concept around equivalence set restrictions is as follows. For an item $j$, we partition the $C$ classes into disjoint equivalence sets $\bigcup_{b=1}^{C} E_{j b}=\mathbb{Z}_{C}$. Classes in the same equivalence set share a common response probabilities $P(x_{ij}|c_{i}=c,\text{Model Parameters})$. Base class matrix $\mymatrix{B}$ can be seen as an index indicating which equivalence set a class belongs to for each item. As defined here equivalence set restrictions can be applied to categorical responses $x_{ij}$. In this paper we provide identifiability conditions for categorical responses, but our simulations and applications focus on Bernoulli responses.

Where a traditional RLAMs apply restrictions to $C=2^{K}$ classes using a $Q$-matrix or $\delta$-matrix, equivalence set restrictions allow for $C \in \mathbb{N}$ classes and offer more generic restrictions. For examples of these more general restrictions see Table~\ref{table:restrictionCount} and Table~\ref{table:esRestrictions4}. Table~\ref{table:restrictionCount} shows that equivalence set restrictions offer a larger number of possible restrictions. For example, the cardinality of the equivalence set restrictions are described by the Bell number, $\mathbb B_C$, which counts the number of ways to partition a set with exactly $C$ elements and satisfies the recurrence relation $\mathbb B_{n+1} = \sum_{k=0}^n \binom{n}{k} \mathbb B_k$. When $C$ is of the form $2^K$ our equivalence set approach allows for $\mathbb B_{2^{K}}$ possible number of restrictions, which is larger than the $2^K$ and $2^{2^K-1}$ restrictions available from the $\mymatrix{Q}$ and $\mymatrix{\delta}$ frameworks. Furthermore, the equivalence set restrictions are valid for $C \in \mathbb{N}$ whereas the $\mymatrix{Q}$ and $\mymatrix{\delta}$ restrictions are only valid when $C=2^K$ for $K\in\mathbb N$. 

Table~\ref{table:esRestrictions4} shows the different types of restrictions possible on each model when $C=4$ (with $K=2$). We see specific examples of restrictions possible under equivalence sets, but not under $\mymatrix{Q}$-matrix or $\mymatrix{\delta}$-matrix restrictions. For instance, the $\mymatrix{Q}$-matrix or $\mymatrix{\delta}$-matrix restrictions are unavailable for eight of the $\mathbb B_4=15$ ways to form equivalence sets. In contrast, there is only one type of $\mymatrix{\delta}$ restriction with $\myvec{\delta}_j=[1,1,1,0]$, which corresponds with a model in Equation \ref{eq:deltagdm} that includes an intercept, main-effects for $\alpha_1$ and $\alpha_2$, but not two-way interaction term $\alpha_1\alpha_2$, that cannot be produced by the equivalence set formulation. 

We call $B_{c j}$ a base class, and call $\myvec{B}_{j}:=[B_{1 j},\cdots,B_{C j}]^{\top}$ a base class vector. By convention the base class vector starts at $\delta_{1 j}=1$, and the first appearance of each new base class increments by one: $\delta_{b j}=\max\{\delta_{1 j},\cdots, \delta_{b-1,j}\}+1$ when $\delta_{b j} \notin \{\delta_{1 j},\cdots, \delta_{b-1,j}\}$. Additionally $\nbaseclass_{j} := |\{B_{c j} : c\}|$ denotes the number of base classes under item $j$.

\begin{table}[!htbp]
\caption{Number of possible restrictions on a single item for the $\mymatrix{Q}$-Matrix, $\mymatrix{\delta}$-matrix, and Equivalence Set  restrictions. }
\formattable{
\begin{tabular}{cccc}
Classes
  (C) & \makecell{$\mymatrix{Q}$-Matrix} & \makecell{$\mymatrix{\delta}$-Matrix} & \makecell{Equivalence Set} \\\hline
$2^{1}$ & 2 & 2 & 2 \\
$2^{2}$ & 4 & 8 & 15 \\
$2^{3}$ & 8 & 128 & 4,140 \\
$2^{4}$ & 16 & 32,768
 & $>10^{10}$ \\
 $2^{K}$ & $2^{K}$ & $2^{(2^{K}-1)}$
 & $\mathbb B_{2^{K}}$ \\
 $C \neq 2^{K}$ & - & - & $\mathbb B_{c}$\\\hline 
\end{tabular}
}
\caption*{\subheader{Note} The Bell number $\mathbb B_{C}$ gives the number of ways $C$ objects can be partitioned \citep{Abramowitz1965}.}
\label{table:restrictionCount}
\end{table}

\begin{table}[!hbtp]
\caption{All possible restrictions for $C=4$ classes under equivalence set restrictions. }
\formattable{
\begin{tabular}{l|cccc|c|c}
 & \multicolumn{4}{c|}{Base Classes Vector $\myvec{B}_{j}$} 
 & \makecell[c]{Possible Under\\ $\myvec{Q}$ restrictions?}
 & \makecell[c]{Possible Under\\ $\myvec{\delta}$ restrictions?}
 \\
Equivalence Sets $E_{j b}$                   & $B_{j,c=1}$ & $B_{j 2}$ & $B_{j 3}$ & $B_{j 4}$ &  \\\hline
\{ 1, 2, 3, 4   \}                 & 1 & 1 & 1 & 1& $\myvec{Q}_{j}=[0,0]$ & $\myvec{\delta}_{j}=[1,0,0,0]$ \\
\{ 1, 2, 3 \}, \{ 4 \}             & 1 & 1 & 1 & 2& No                    & $\myvec{\delta}_{j}=[1,0,0,1]$ \\
\{ 1, 2, 4 \}, \{ 3 \}             & 1 & 1 & 2 & 1& No                    & No                           \\
\{ 1, 3, 4 \}, \{ 2 \}             & 1 & 2 & 1 & 1& No                    & No                           \\
\{ 1 \}, \{ 2, 3, 4 \}             & 1 & 2 & 2 & 2& No                    & No                           \\
\{ 1, 2 \}, \{ 3, 4 \}             & 1 & 1 & 2 & 2& $\myvec{Q}_{j}=[0,1]$ & $\myvec{\delta}_{j}=[1,0,1,0]$ \\
\{ 1, 3 \}, \{ 2, 4 \}             & 1 & 2 & 1 & 2& $\myvec{Q}_{j}=[1,0]$ & $\myvec{\delta}_{j}=[1,1,0,0]$ \\
\{ 1, 4 \}, \{ 2, 3 \}             & 1 & 2 & 2 & 1& No                    & No                           \\
\{ 1, 2 \}, \{ 3 \}, \{ 4 \}       & 1 & 1 & 2 & 3& No                    & $\myvec{\delta}_{j}=[1,0,1,1]$ \\
\{ 1, 3 \}, \{ 2 \}, \{ 4 \}       & 1 & 2 & 1 & 3& No                    & $\myvec{\delta}_{j}=[1,1,0,1]$ \\
\{ 1, 4 \}, \{ 2 \}, \{ 3 \}       & 1 & 2 & 3 & 1& No                    & No                           \\
\{ 1 \}, \{ 2, 3 \}, \{ 4 \}       & 1 & 2 & 2 & 3& No                    & No                           \\
\{ 1 \}, \{ 2, 4 \}, \{ 3 \}       & 1 & 2 & 3 & 2& No                    & No                           \\
\{ 1 \}, \{ 2 \}, \{ 3, 4 \}       & 1 & 2 & 3 & 3& No                    & No                           \\
\{ 1 \}, \{ 2 \}, \{ 3 \}, \{ 4 \} & 1 & 2 & 3 & 4& $\myvec{Q}_{j}=[1,1]$ & $\myvec{\delta}_{j}=[1,1,1]$ \\
\makecell[c]{No} &  \multicolumn{4}{c|}{No}  & No & $\myvec{\delta}_{j}=[1,1,0]$\\ \hline
\end{tabular}
}
\subheader{Note} The mapping between class and latent traits is as follows: Classes $0,1,2,3$ have latent traits $[0,0],[1,0], [0,1], [1,1]$ respectively.
\label{table:esRestrictions4}
\end{table}

\FloatBarrier

%% file: PaperSections/0250_identifiability.tex
\FloatBarrier
\section{Identifiability}

In this section we establish conditions for generic identifiability of equivalence set RLCMs. We start with some definitions.

\begin{definition}
We say $\mymatrix{B} \preceq \tilde{\mymatrix{B}}$ if for every $j,b$ there exists a $b'$ such that $E_{j b} \subseteq \tilde{E}_{j b'}$. If $\mymatrix{B} \preceq \tilde{\mymatrix{B}}$ then we say $\tilde{\mymatrix{B}}$ is a merged base class of $\mymatrix{B}$.
\end{definition}

\begin{theorem}
For base class matrices $\mymatrix{B}$ and $\tilde{\mymatrix{B}}$, the following are equivalent:
\begin{enumerate}
\item Matrix $\tilde{\mymatrix{B}}$ is a merged base class matrix of $\mymatrix{B}$.
\item For every $j,b'$ there exists indices $S_{j b'} \subseteq \mathbb{Z}_{C}$ such that $\tilde{E}_{j b'} = \bigcup_{b \in S_{j b'}} E_{j b}$.
\item There exists a transformation $g:(\mathbb{Z}_{J}\times\mathbb{Z}_{C})\rightarrow \mathbb{Z}_{C}$ such that $\tilde{B}_{cj} := g(j,B_{cj})$.
\end{enumerate}
\end{theorem}

\begin{example}
\begin{align*}
\mymatrix{B} 
=
\left[\begin{array}{lll}
1 & 1 & 1 \\
1 & 2 & 1 \\
2 & 1 & 1 \\
2 & 3 & 2 \\
3 & 1 & 2 \\
3 & 4 & 2
\end{array}\right]
\preceq
\left[\begin{array}{lll}
1 & 1 & 1 \\
1 & 2 & 1 \\
1 & 1 & 1 \\
1 & 3 & 2 \\
2 & 1 & 2 \\
2 & 2 & 2
\end{array}\right]
= \tilde{\mymatrix{B}}
\end{align*}
\end{example}

\begin{restatable}[Generic Identifiability of Equivalence Set RLCM]{theorem}{thmEsIdentifiability}
\label{thm:EsIdentifiability}
For a fixed choice of base classes $\mymatrix{B}$, an equivalence set RLCM is generically identifiable given the following. There must exist a merged base class matrix $\tilde{\mymatrix{B}}$ and a tripartition of items $\bigcup_{k=1}^{3} \mathcal{J}_{k} = \mathbb{Z}_{J}$ which following the below criteria. For convenience let $\mymatrix{B}^{(k)}:=\mymatrix{B}_{\mathcal{J}_{k}}:=[\myvec{B}_{j}:j\in \mathcal{J}_{k}]\in \mathbb{Z}_{J}^{J \times |\mathcal{J}_{k}|}$ be the matrix formed by the base classes vectors of items $\mathcal{J}_{k}$. In the third tripartition, we require that the rows of $\mymatrix{B}^{(3)}$ are unique. For the first two tripartitions we require the following:
\begin{enumerate}
\item For every item, the number of merged base classes cannot exceed the number of response levels: $\tilde{\nbaseclass}_{j} \leq m_{j}$.
\item Every row of merged base matrix $\tilde{\mymatrix{B}}^{(k)}$ must be unique.
\item The number of possible response patterns must meet or exceed the number of classes: $\prod_{j \in \mathcal{J}_{k}} m_{j} \geq C$.
\end{enumerate}
\end{restatable}

\begin{proof}
See Appendix~\ref{appendix:esRestrictions}.
\end{proof}

\begin{example}
Suppose we have $J=6$ items where the first two items have $m_{j}=3$ levels, and the remaining items have $m_{j}=2$ levels. With $C=5$ classes our restrictions $\mymatrix{B}$ are:

\begin{align*}
\mymatrix{B} 
& := \left[\begin{tabular}{llllll}
1 & 1 & 1 & 1 & 1 & 1 \\
2 & 2 & 2 & 2 & 1 & 2 \\
3 & 3 & 3 & 1 & 1 & 3 \\
2 & 1 & 4 & 3 & 2 & 3 \\
1 & 4 & 4 & 2 & 3 & 3
\end{tabular}\right]
\end{align*}

Under this choice of $\mymatrix{B}$, we will demonstrate that the equivalence set RLCM has generic identifiability. Take tripartition of items $\mathcal{J}_{1}=\{1,4\}$, $\mathcal{J}_{2}=\{2,3\}$, and $\mathcal{J}_{3}=\{5,6\}$. Note that for the first two tripartitions the number of possible responses ($6$) exceeds the number of classes ($5$). Below we have the original tripartioned and then merged base class matrices. 
\begin{align*}
\mymatrix{B}^{(1)} 
& := \begin{blockarray}{ll}
\myvec{B}_{1} & \myvec{B}_{4} \\
\begin{block}{[ll]}
1 & 1 \\
2 & 2 \\
3 & 1 \\
2 & 3 \\
1 & 2 \\
\end{block}
\end{blockarray}
& \mymatrix{B}^{(2)} 
& := \begin{blockarray}{ll}
\myvec{B}_{2} & \myvec{B}_{3} \\
\begin{block}{[ll]}
1 & 1 \\
2 & 2 \\
3 & 3 \\
1 & 4 \\
4 & 4 \\
\end{block}\end{blockarray}
& \mymatrix{B}^{(3)}
& := \begin{blockarray}{ll}
\myvec{B}_{5} & \myvec{B}_{6} \\
\begin{block}{[ll]}
1 & 1 \\
1 & 2 \\
1 & 3 \\
2 & 3 \\
3 & 3 \\
\end{block}\end{blockarray}
\\
\tilde{\mymatrix{B}}^{(1)} 
& := \left[\begin{array}{ll}
1 & 1 \\
2 & 2 \\
3 & 1 \\
2 & 1 \\
1 & 2 \\
\end{array}\right]
& \tilde{\mymatrix{B}}^{(2)} 
& := \left[\begin{array}{ll}
1 & 1 \\
2 & 1 \\
3 & 1 \\
1 & 2 \\
3 & 2 \\
\end{array}\right]
& \tilde{\mymatrix{B}}^{(3)} 
& := \mymatrix{B}^{(3)}
\end{align*}

Note that the columns of the first two merged base class matrices have no more base classes than response levels to an item. Additionally note that the rows of all three merged base class matrices are unique. By Theorem~\ref{thm:EsIdentifiability} we have generic identifiability.

\end{example}

These identifiability conditions can be checked by way of a greedy algorithm. Starting with three empty partitions, items are added one at a time, and a corresponding $\tilde{\myvec{B}}_{j}$ is generated. If the algorithm ultimately finds a solution for which identifiability holds, then we have shown identifiability. If no solution is found, we have no evidence supporting identifiability, but it is possible that a more exhaustive search could show identifiability. Such a method is described in Appendix~\ref{appendix:esRestrictions}. 

Theorem~\ref{thm:EsIdentifiability} can be applied to traditional Q-matrix RLCMs. Corollary~\ref{thm:Qidentifiability} below is equivalent to the results shown by \cite{gu2018}. Additionally Theorem~\ref{thm:EsIdentifiability} can be further extended to Q-matrices on non-binary categorical responses.

\begin{restatable}{corollary}{thmQidentifiability}
\label{thm:Qidentifiability}
For a Q-matrix RLCM on binary responses, generic identifiability holds for a fixed Q-matrix given the following. Up to relabeling of items, the Q-matrix can be put into block diagonal form $\mymatrix{Q}=[\mymatrix{M}_{1}^{\top},\mymatrix{M}_{2}^{\top},\mymatrix{M}_{3}^{\top}]^{\top}$. Matrices $\mymatrix{M}_{1}$ and $\mymatrix{M}_{2}$ must be $C\times C$ matrices with diagonal elements of $1$. For $\mymatrix{M}_{3}$ each column must include one or more $1$ values.
\end{restatable}

%% file: PaperSections/0300_dependentBeta.tex
\FloatBarrier
\section{Repelled Beta Distribution}\label{section:dependentBeta}

In this section we introduce a a novel prior called the repelled beta distribution. This prior is designed to encourage well separated class response probabilities. In general when base classes differ $B_{j c} \neq B_{j c'}$, we prefer that $P(x_{ij}|c_{i}=c,\modelParams) \neq P(x_{ij}|c_{i}=c',\modelParams)$ in some meaningful way.

\begin{definition}
Let $\myvec{\rho} \in (0,1)^{M}$ follow the repelled beta distribution. Using subscript $\oo{k}$ to denote the $k$'th order statistic, we say $\myvec{\rho} \sim \text{RepelledBeta}(\mymatrix{\alpha}, v)$ is distributed as a repelled beta distribution with density given by:
\begin{align}
P(\myvec{\rho})
&\propto \prod_{k=1}^{M} [\rho_{k}^{\alpha_{k 1}-1}(1-\rho_{k})^{\alpha_{k 2}-1}] \prod_{k=2}^{M} (\rho_{\oo{k}}-\rho_{\oo{k-1}})^{v}
\label{eq:depThetaJoint}
\end{align}
on the region where $0 < \rho_{k} < 1$ for all $k$.
\end{definition}

\begin{figure}[!hbtp]
\centering
\includegraphics[max width=\textwidth]{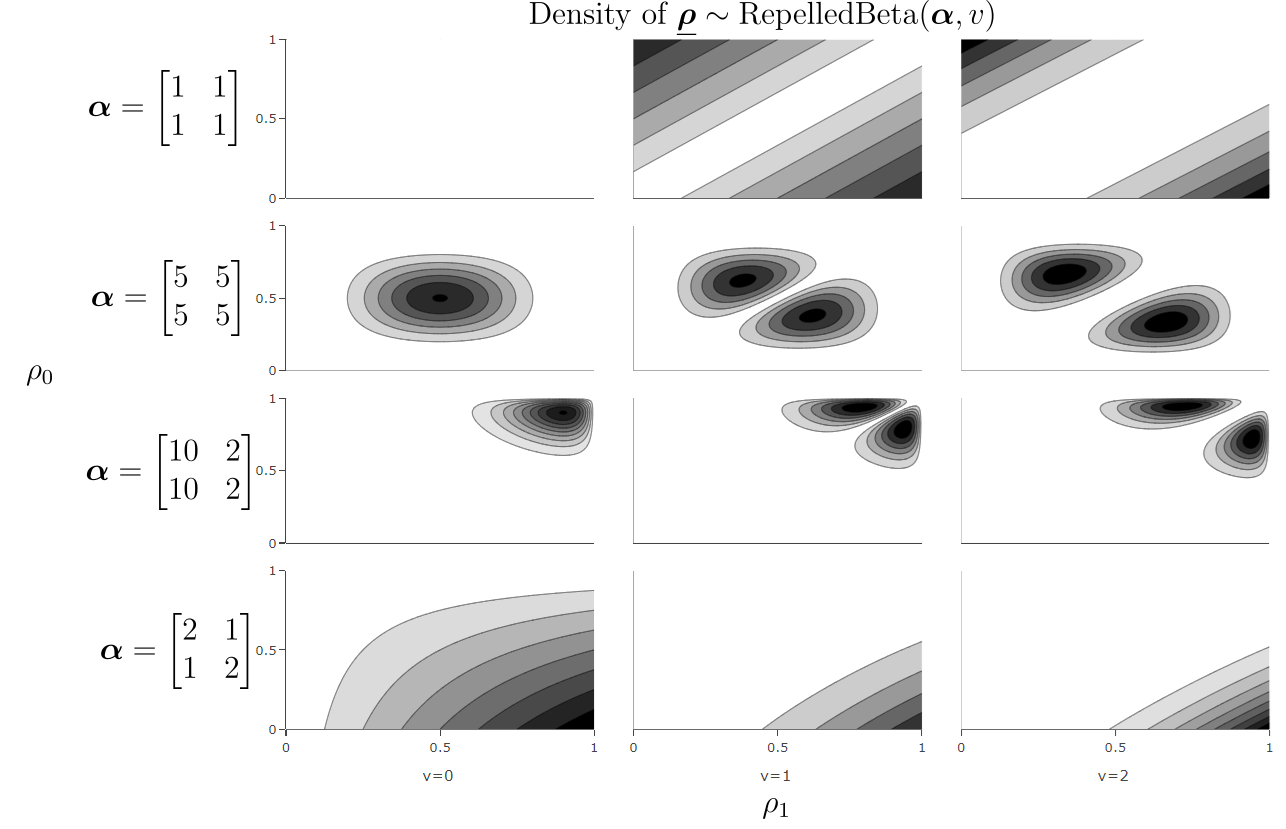}
\caption{Density of repelled beta distribution. Each row of matrix $\mymatrix{\alpha}$ corresponds to a different component of $\myvec{\rho}$.}
\label{fig:repelledBetaDensity}
\end{figure}

\begin{remark}
For $k\neq k'$, the repelled beta distribution is designed to encourage $|\rho_{k}-\rho_{k'}|\gg 0$. In particular when $v>0$, the pdf approaches 0 as two rhos approach each other: $P(\myvec{\rho})\to 0$ as $\rho_j\to \rho_k$ for all $j\neq k$. When $v=0$, the repelled beta distribution reduces to the product of $M$ independent beta random variables. As $v$ gets large, the elements of $\myvec{\rho}$ get increasingly dependent and tend further apart. In the 2-dimensional case, the repelled beta is illustrated in Figure~\ref{fig:repelledBetaDensity}.
\end{remark}

\begin{restatable}{theorem}{thmdepThetaPrior}
\label{thm:depThetaPrior}
When $\alpha_{k 1}=\alpha_{k 2}=1$ for all $k$ then the repelled beta distribution has density:
\begin{align}
P(\rho_{1},\cdots,\rho_{M})
&= \frac{\Gamma((M-1)(v+1)+2)}{M!\Gamma^{M-1}(v+1)} \prod_{k=2}^{M} (\rho_{\oo{k}}-\rho_{\oo{k-1}})^{v}
\label{eq:depThetaPrior}
\end{align}
\end{restatable}

\begin{proof}
When $\alpha_{k 1}=\alpha_{k 2}=1$, the differences $\rho_{\oo{k}}-\rho_{\oo{k-1}}$ follow a Dirichlet distribution. This is used to find the normalizing constant. For details see Appendix~\ref{appendix:repelledBeta}.
\end{proof}

\begin{restatable}{theorem}{thmdepThetaConjugacyb}\label{thm:thmdepThetaConjugacyb}
The Repelled Beta distribution is conjugate on Bernouli responses. Let $\myvec{\rho} \sim \text{RepelledBeta}(\mymatrix{\alpha}, v)$. Take responses $z_{i k} \sim \text{Bernouli}(\rho_{k})$ for $i \in \{1, \cdots, l_{k}\}$.
\begin{align*}
\myvec{\rho}|\mymatrix{Z}
& \sim \text{RepelledBeta}(\mymatrix{\alpha}+\mymatrix{n}, v) \\
n_{k 1} & := \sum_{i=1}^{l_{k}} z_{i k}
;\quad n_{k 2} := l_{k} - n_{k 1}
\end{align*}
\end{restatable}

\begin{proof}
Proof is analogous to the conjugacy of beta and Bernouli distributions.
\end{proof}

\begin{remark}[Sampling]
We sample from the repelled beta distribution using a rejection sampler. The rejection sampler proposes using independent beta distributions.
\end{remark}

%% file: PaperSections/0500_posteriors.tex
\FloatBarrier
\section{Posterior Approximation}

In this section we detail the priors, full conditionals, and Monte Carlo Markov Chain (MCMC) sampling used by equivalence set RLCMs.

The class response probabilities of item $j$ under class $c$ is denoted $\theta_{c j} := P(x_{ij}|c_{i}=c,\theta_{c j})$. Classes belonging to the same base class share the same response probabilities. We let $\theta'_{b j} := P(x_{ij}|c_{i}\in E_{j b},\theta'_{b j})$ denote the response probabilities for base class $b$ in item $j$. As described in Section~\ref{section:eqsets}, let $\myvec{B}_{j}$ denote a base class vector, and $\nbaseclass_{j}$ denote the number of base classes. Let parameter $\myvec{\eta}_{j}$, outlined in Definition~\ref{def:priors}, partially control the number of base classes. The distribution of $\myvec{\theta}_{j}$ is a mixture over the different choices of base classes $\myvec{\beta}_{j}$:
\begin{align*}
P(\myvec{\theta}_{j}\mid \myvec{\eta}_{j},v) 
&\propto \sum_{\nbaseclass_j}\sum_{\myvec{B}_j} I(\myvec{\theta}_{j}\equiv \myvec{\theta}_{j}'|\myvec{B}_j)P(\myvec{\theta}_{j}'\mid \nbaseclass_{j},v)P(\myvec{B}_j\mid \nbaseclass_{j}) P(\nbaseclass_j\mid \myvec{\eta}_{j})\\
I(\myvec{\theta}_{j}\equiv \myvec{\theta}_{j}'|\myvec{B}_j)
& := \prod_{c} I(\theta_{c j} = \theta_{B_{c j}j}')
\end{align*}

\begin{figure}[!hbtp]
\centering
\includegraphics[max width=0.8\textwidth]{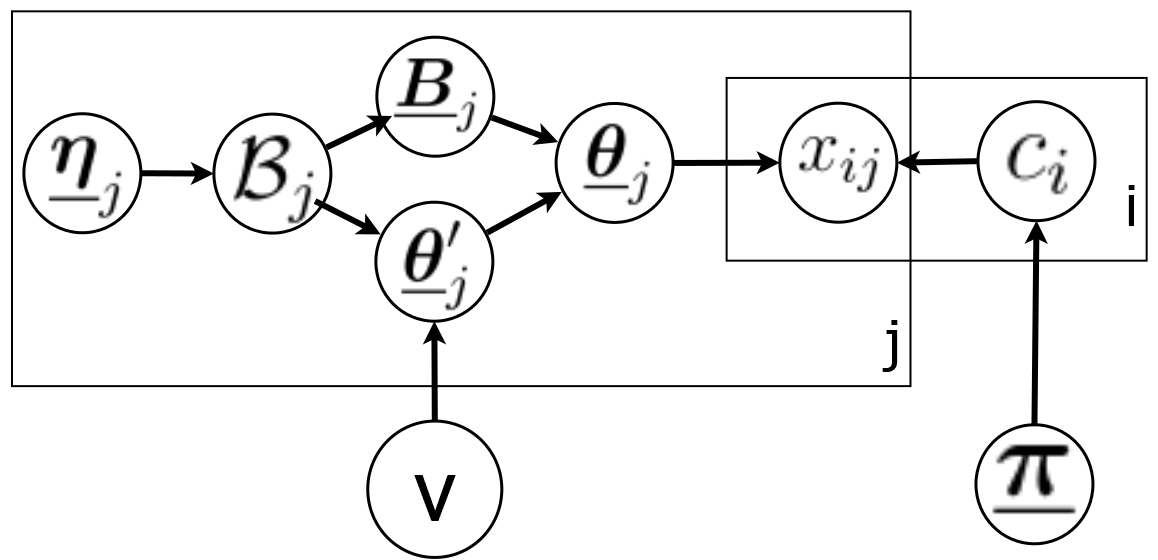}
\caption{Dependencies between parameters in equivalence set RLCMs.}
\label{fig:dependencyDiagram}
\end{figure}

The relationship between our parameters is illustrated in Figure~\ref{fig:dependencyDiagram} with the specific priors given in Definition~\ref{def:priors}. Our full joint distribution is:
\begin{align*}
&P(\myvec{\pi},v,\mymatrix{\eta},\myvec{\nbaseclass},\mymatrix{B},\mymatrix{\theta}',\mymatrix{\theta},\myvec{c},\mymatrix{x})\\
&= P(\myvec{\pi})P(v) \prod_{j} \left[P(\myvec{\eta}_{j}) P(\nbaseclass_{j}|\myvec{\eta}_{j})P(\myvec{B}_{j}|\nbaseclass_{j})P(\myvec{\theta}'_{j}|v,\nbaseclass_{j})I(\myvec{\theta}_{j}\equiv \myvec{\theta}_{j}'|\myvec{B}_j)\right] \prod_{i} \left[P(c_{i}|\myvec{\pi}) P(\myvec{x}_{i}|\theta,c_{i})\right].
\end{align*}

\begin{definition}[Priors]\label{def:priors} Below we give the prior of each parameter in the equivalence set RLCM.\\\\
\formattable{
\begin{tabular}{lll}
Description & Variable & Distribution \\ \hline
Stick Breaking Probability: & $\eta_{k j}$ & $\sim \text{Beta}(\myvec{\alpha}_{k j}^{(\eta)})$ \\
Number of Base Classes: & $\nbaseclass_{j}|\myvec{\eta}_{j}$ & $\sim \text{StickBreaking}(\myvec{\eta}_{j})$ \\
& & $= (\prod_{k}^{\nbaseclass_{j}-1} \eta_{k j}) (1-\eta_{\nbaseclass_{j} j})^{I(\nbaseclass_{j}<C)}$ \\
Base Classes: & $P(\myvec{B}_{j}|\nbaseclass_{j})$ & $= \frac{1}{\stirling{C}{\nbaseclass_{j}}}\propto 1$ \\
Repulsion: & $P(v)$ & $\propto v^{d_{1}} e^{d_{2} v} I(0 < v < \text{MaxV})$ \\
Base Class Response Probabilities: & $(\myvec{\theta}_{j}'|\nbaseclass_{j},v)$ & $\sim \text{RepelledBeta}(\mymatrix{\alpha}=\mymatrix{1}, v)$ \\
Class response probabilities: & $(\theta_{jc}|\myvec{\theta}_{j}',\myvec{B}_{j})$ & $= \theta_{j B_{c j}}'$ \\
Class Prior: & $\myvec{\pi}$ & $\sim \text{Dirichlet}(\myvec{\alpha}^{(c)})$ \\
Classes: & $c_{i}$ & $\sim \text{Categorical}(\myvec{\pi})$ \\
Responses: & $(x_{ij}|\myvec{\theta}_{j},c_{i})$ & $\sim \text{Bernouli}(\theta_{c_{i} j})$
\end{tabular}
}
Above $\stirling{m}{k}$ refers to the Stirling number of the second kind, denoting the number of possible ways to partition $m$ objects into exactly $k$ nonempty sets. For prior on $v$, parameters $d_{1},d_{2}>0$ are positive to ensure that the greatest density given to the largest values of $v$, inducing regularization. By default $d_{1}=d_{2}=1$ and $\text{MaxV}=2$.
\end{definition}

Our MCMC update steps are outlined in Process~\ref{def:mcmc}. During these steps we collapse on the stickbreaking parameters $\myvec{\eta}_{j}$. We can safely marginalize on $\myvec{\eta}_{j}$ noting that:

\begin{theorem}\label{thm:etaDoesNotMatter}
For any $\myvec{\zeta} \in \text{Simplex}_{C}$, there exists an $\mymatrix{\alpha}_{j}^{(\eta)}$ such that the prior of our base classes collapsed on $\myvec{\eta}_{j}$ are given by: \begin{align}
\nbaseclass_{j}
&\sim \text{Categorical}(\myvec{\zeta}) \\
P(\myvec{B}_{j}) 
&= \frac{\zeta_{\nbaseclass_{j}}}{\stirling{C}{\nbaseclass_{j}}} \label{eq:basevecProptoProb}
\end{align}
with $\stirling{M}{K}$ denoting the Stirling number of the second kind.
\end{theorem}

\begin{proof}
Proof in Appendix~\ref{appendix:posteriorsStickbreaking}.
\end{proof}

\begin{remark}\label{remark:q}
By default in our simulations and applications, we choose $\mymatrix{\alpha}_{j}^{(\eta)}$ such that $P(\myvec{B}_{j}) \propto \lambda^{\nbaseclass_{j}}$ for hyperparameter $\lambda \in (0,1]$. This mirrors the truncated geometric distribution. The code supports any general $\mymatrix{\alpha}_{j}^{(\eta)}$ and any general $\myvec{\zeta}$ as defined in Theorem~\ref{thm:etaDoesNotMatter}.
\end{remark}

\begin{process}[MCMC]\label{def:mcmc} We give the Monte Carlo Markov Chain steps used to fit ESRLCMs. For details see Appendix~\ref{appendix:posteriors}. \\\\
\formattable{
\begin{tabular}{llp{7cm}}
Variable & Method & Details \\ \hline
$\myvec{B}_{j},\theta'_{j}|v,\myvec{x}_{j}$ & Reversible Jump & Proposal: A random class has $B_{c j}$ updated. Original and modified base class are given a new $\theta'_{b j}$. \\
$v|\mymatrix{\theta}'$ & Metropolis & Proposal: Triangular distribution with mode at maximum a-posterior value.\\
$\myvec{\theta}_{j}'|\myvec{B}_{j},v,\myvec{x}_{j}$ & Gibbs & $\sim \text{RepelledBeta}(\mymatrix{1}+\mymatrix{n}_{j}^{(\theta)}, v)$
\\ 
& & $n^{(\theta)}_{j b r}:= \sum_{i} I(B_{c_{i} j}=b,x_{i j}=1-r)$ \\
$\myvec{\pi}|\myvec{c}$ & Gibbs & $\sim \text{Dirichlet}(\myvec{\alpha}^{(c)}+\myvec{n}^{(c)})$ \\
& & $n^{(c)}_{c'}:=\sum_{i} I(c_{i}=c')$ \\
$c_{i}|\myvec{x}_{i},\myvec{\pi},\mymatrix{\theta}$ & Gibbs & $\sim \text{Categorical}\left(\frac{\pi_{c}P(\myvec{x}_{i}|c_{i}=c,\mymatrix{\theta})}{\sum_{c'} \pi_{c}P(\myvec{x}_{i}|c_{i}=c',\mymatrix{\theta})}\right)$
\end{tabular}
}
\end{process}

%% file: PaperSections/0600_simulations.tex
\FloatBarrier
\section{Simulation Studies}\label{section:sims}

We conducted Monte Carlo simulation studies to verify the accuracy of the equivalence set RLCMs. Here, we will generate random datasets following a specific distribution. We then fit ESRLCMs and several competitors on each dataset. For each model we examine the recovery of restrictions, and the goodness of fit of the models.

The simulated datasets are generated under an ESRLCM model with $J=32$ Bernouli items. Each item has $\nbaseclass_{j} \in \{2,\cdots,8\}$ number of base classes with restrictions given in Appendix~\ref{appendix:simulations}. Each base class has response probabilities evenly spaced between $\theta_{b j}' = 1/(2 \nbaseclass_{j})$ and $\theta_{b j}' = 1-1/(2 \nbaseclass_{j})$ inclusive. We fit each combination of the following:
\begin{itemize}
\item Data:
    \begin{itemize}
    \item Number of Classes: Data is generated with differing numbers of classes $C \in \{4,5,8,11,16\}$
    \item Sample Sizes: Data has sample size in $n \in \{500, 1000, 2000, 4000, 8000\}$.
    \end{itemize}
\item Models:
    \begin{itemize}
    \item ESRLCM. We fit four types of ESRLCMs based on each combination of:
    \begin{itemize}
        \item $v$: Response probabilities $\theta'$ have regularizing parameter $v$ either fixed to zero, or $v>0$ and allowed to vary.
        \item $\lambda$: The number of base classes $\nbaseclass_{j}$ has regularizing parameter $\lambda$ as defined in Remark~\ref{remark:q}. This takes values $\lambda \in \{0.5, 1\}$.
    \end{itemize}
    \item Regularized LCM, Frequentist. We fit a regularized LCM based on \cite{Chen2016}. We use the reglca function available in the R CDM package \citep{jsscdm}. The model is fitted with a SCAD penalty with $\lambda=0.1$.
    \item Unrestricted LCM. We fit a traditional LCM without restrictions.
    \item RLAM, Delta-Matrix. We fit a restricted latent attribute model with $\delta$-matrix restrictions \citep{egdmdirac}. This model assumes monotonicity; higher levels of class attributes $\myvec{\alpha}_{i}$ correspond with higher response probabilities. We only fit this model when $C \in \{4,8,16\}$, matching model assumptions.
    \end{itemize}
\end{itemize}
A total of $200$ datasets are generated as described above. Models are fit in a single chain with $5,000$ warm-up iterations followed by $5,000$ main iterations.

Performance is measured using $20,000$ out of sample datapoints, generated alongside each dataset. In Table~\ref{table:simGof}, we see that ESRLCMs with $v>0$ tend to outperform ESRLCMs with $v=0$. When compared, equivalence set RLCMs outperform the other models. Restricted latent attribute models perform worst due to model misspecification, including lack of monotonicity.

We measure recovery of restrictions by looking at each pair of classes under each item. Each pair of response probabilities are either the same or distinct. \quote{Restriction sensitivity} looks at every pair which, in truth, has identical response probabilities ($\theta_{c j}=\theta_{c' j}$), and reports what percent of these pairs were modeled as having the same response probabilities. \quote{Restriction Specificity} conversely looks at each pair which, in truth, have differing response probabilities ($\theta_{c j}\neq\theta_{c' j}$), and reports what percent of pairs were modeled as having distinct response probabilities. In Table~\ref{table:simsRestrictions}, we see that ESRLCMs do a good job of recovering restrictions, even when the number of classes is large.

\begin{table}[]
\formattable{
\begin{tabular}{c|lll|rrrrr|rrrrr}
          & \multicolumn{3}{c|}{Model}                             & \multicolumn{5}{|l}{\makecell{\small{Percent of Times a Model}\\\small{Had best OOS Predictions}}} & \multicolumn{5}{|c}{OOS LogLikelihood, Average} \\
&&&&&&&&&&&&& \\
&&&&&&&&&&&&& \\
\makecell{$C$} & Model                   & v                & $\lambda$   & \myrotate{c}{55}{n=500}        & \myrotate{c}{55}{1,000}        & \myrotate{c}{55}{2,000}       & \myrotate{c}{55}{4,000}       & \myrotate{c}{55}{8,000}       & n=500    & 1,000   & 2,000   & 4,000   & 8,000   \\\hline
4  & ESRLCM       & $v=0$ & $\lambda=0.5$      & 0& 0& 0& 4& 7& -18.779& -18.716& -18.688& -18.675& -18.671\\
   &              & $v=0$ & $\lambda=1$        & 0& 0& 0& 1& 4& -18.776& -18.717& -18.688& -18.675& -18.671\\
   &              & $v>0$ & $\lambda=0.5$      & 18& 72& 77& 55& 48& -18.771& -18.713& -18.687& -18.675& -18.671\\
   &              & $v>0$ & $\lambda=1$        & 80& 26& 20& 33& 30& -18.768& -18.714& -18.687& -18.675& -18.671\\
   & \multicolumn{3}{l|}{Regularized LCM} & 0& 0& 2& 7& 10& -18.893& -18.784& -18.724& -18.704& -18.692\\
   & \multicolumn{3}{l|}{Unrestricted LCM}             & 0& 0& 0& 0& 0& -18.794& -18.73& -18.696& -18.679& -18.673\\
   & \multicolumn{3}{l|}{RLAM, Delta-matrix}           & 0& 0& 0& 0& 0& -20.956& -20.922& -20.899& -20.892& -20.888\\
&&&&&&&&&&&&& \\
5  & ESRLCM       & $v=0$ & $\lambda=0.5$      & 0& 0& 1& 4& 6& -18.994& -18.92& -18.892& -18.878& -18.873\\
   &              & $v=0$ & $\lambda=1$        & 0& 0& 0& 0& 4& -18.992& -18.922& -18.893& -18.879& -18.873\\
   &              & $v>0$ & $\lambda=0.5$      & 48& 92& 95& 79& 64& -18.984& -18.917& -18.891& -18.878& -18.872\\
   &              & $v>0$ & $\lambda=1$        & 52& 8& 4& 18& 26& -18.984& -18.918& -18.892& -18.878& -18.872\\
   & \multicolumn{3}{l|}{Regularized LCM} & 0& 0& 0& 0& 0& -19.219& -19.129& -19.017& -18.979& -18.937\\
   & \multicolumn{3}{l|}{Unrestricted LCM}             & 0& 0& 0& 0& 0& -19.027& -18.948& -18.907& -18.886& -18.876\\
&&&&&&&&&&&&& \\
8  & ESRLCM       & $v=0$ & $\lambda=0.5$      & 0& 0& 0& 0& 0& -19.074& -18.928& -18.852& -18.813& -18.802\\
   &              & $v=0$ & $\lambda=1$        & 0& 0& 0& 0& 0& -19.060& -18.920& -18.849& -18.813& -18.803\\
   &              & $v>0$ & $\lambda=0.5$      & 0& 0& 0& 16& 78& -19.054& -18.917& -18.847& -18.811& -18.802\\
   &              & $v>0$ & $\lambda=1$        & 64& 99& 100& 84& 22& -19.042& -18.909& -18.843& -18.811& -18.802\\
   & \multicolumn{3}{l|}{Regularized LCM} & 0& 0& 0& 0& 0& -20.041& -19.891& -19.724& -19.632& -19.565\\
   & \multicolumn{3}{l|}{Unrestricted LCM}             & 36& 1& 0& 0& 0& -19.046& -18.922& -18.858& -18.822& -18.808\\
   & \multicolumn{3}{l|}{RLAM, Delta-matrix}           & 0& 0& 0& 0& 0& -21.163& -21.102& -21.070& -21.048& -21.044\\
&&&&&&&&&&&&& \\
11 & ESRLCM       & $v=0$ & $\lambda=0.5$      & 0& 0& 0& 0& 0& -19.391& -19.210& -19.119& -19.077& -19.063\\
   &              & $v=0$ & $\lambda=1$        & 10& 0& 0& 0& 0& -19.385& -19.206& -19.117& -19.077& -19.063\\
   &              & $v>0$ & $\lambda=0.5$      & 48& 48& 38& 48& 96& -19.379& -19.201& -19.114& -19.074& -19.062\\
   &              & $v>0$ & $\lambda=1$        & 37& 51& 62& 52& 4& -19.380& -19.201& -19.113& -19.074& -19.062\\
   & \multicolumn{3}{l|}{Regularized LCM} & 0& 0& 0& 0& 0& -20.570& -20.382& -20.260& -20.127& -20.022\\
   & \multicolumn{3}{l|}{Unrestricted LCM}             & 4& 0& 0& 0& 0& -19.395& -19.225& -19.138& -19.093& -19.074\\
&&&&&&&&&&&&& \\
16 & ESRLCM       & $v=0$ & $\lambda=0.5$      & 22& 38& 21& 4& 0& -19.815& -19.558& -19.433& -19.373& -19.352\\
   &              & $v=0$ & $\lambda=1$        & 28& 8& 0& 0& 0& -19.813& -19.558& -19.434& -19.375& -19.353\\
   &              & $v>0$ & $\lambda=0.5$      & 35& 52& 79& 96& 100& -19.814& -19.558& -19.432& -19.372& -19.351\\
   &              & $v>0$ & $\lambda=1$        & 8& 2& 0& 0& 0& -19.816& -19.561& -19.435& -19.375& -19.353\\
   & \multicolumn{3}{l|}{Regularized LCM} & 0& 0& 0& 0& 0& -21.176& -20.975& -20.817& -20.66& -20.497\\
   & \multicolumn{3}{l|}{Unrestricted LCM}             & 7& 0& 0& 0& 0& -19.826& -19.592& -19.468& -19.403& -19.371\\
   & \multicolumn{3}{l|}{RLAM, Delta-matrix}           & 0& 0& 0& 0& 0& -21.410& -21.358& -21.330& -21.309& -21.283\\
\end{tabular}
}
\caption{On $200$ simulated datsets, goodness of fit is compared across models. Prediction accuracy was measured based on $n=20,000$ out of sample (OOS) observations. The reported OOS LogLikelihood is averaged across observations and simulations.}
\label{table:simGof}
\end{table}

\begin{table}[]
\formattable{
\begin{tabular}{c|lll|rrrrr|rrrrr}
          & \multicolumn{3}{|c}{Model}                        & \multicolumn{5}{|l}{\makecell{Restriction Sensitivity ($\%$)}} & \multicolumn{5}{|l}{\makecell{Restriction Specificity ($\%$)}} \\
&&&&&&&&&&&&& \\
&&&&&&&&&&&&& \\
\makecell{$C$} & Model      & v                    & $\lambda$       & \myrotate{c}{55}{n=500}           & \myrotate{c}{55}{1,000}           & \myrotate{c}{55}{2,000}           & \myrotate{c}{55}{4,000}           & \myrotate{c}{55}{8,000}          & \myrotate{c}{55}{n=500}            & \myrotate{c}{55}{1,000}            & \myrotate{c}{55}{2,000}            & \myrotate{c}{55}{4,000}            & \myrotate{c}{55}{8,000}            \\\hline
4             & ESRLCM          & $v=0$ & $\lambda=0.5$     & 98& 99& 99& 99& 100& 100& 100& 100& 100& 100\\
              &                 & $v=0$ & $\lambda=1$       & 95& 97& 98& 98& 99& 100& 100& 100& 100& 100\\
              &                 & $v>0$ & $\lambda=0.5$     & 99& 99& 100& 100& 100& 99& 100& 100& 100& 100\\
              &                 & $v>0$ & $\lambda=1$       & 97& 99& 99& 100& 100& 100& 100& 100& 100& 100\\
              & \multicolumn{3}{l|}{Regularized LCM} & 95& 99& 99& 99& 99& 97& 99& 99& 99& 99\\
              & \multicolumn{3}{l|}{Unrestricted LCM}             & 0& 0& 0& 0& 0& 100& 100& 100& 100& 100\\
              & \multicolumn{3}{l|}{RLAM, Delta-matrix}           & 22& 21& 21& 22& 20& 77& 78& 78& 77& 77\\
&&&&&&&&&&&&& \\
5             & ESRLCM          & $v=0$ & $\lambda=0.5$     & 97& 98& 99& 99& 99& 99& 100& 100& 100& 100\\
              &                 & $v=0$ & $\lambda=1$       & 93& 96& 97& 98& 98& 100& 100& 100& 100& 100\\
              &                 & $v>0$ & $\lambda=0.5$     & 98& 99& 100& 100& 100& 99& 100& 100& 100& 100\\
              &                 & $v>0$ & $\lambda=1$       & 96& 98& 99& 99& 100& 100& 100& 100& 100& 100\\
              & \multicolumn{3}{l|}{Regularized LCM} & 93& 97& 99& 99& 99& 94& 95& 97& 98& 99\\
              & \multicolumn{3}{l|}{Unrestricted LCM}             & 0& 0& 0& 0& 0& 100& 100& 100& 100& 100\\
&&&&&&&&&&&&& \\
8             & ESRLCM          & $v=0$ & $\lambda=0.5$     & 92& 95& 97& 99& 99& 92& 96& 99& 100& 100\\
              &                 & $v=0$ & $\lambda=1$       & 88& 92& 95& 97& 98& 93& 97& 99& 100& 100\\
              &                 & $v>0$ & $\lambda=0.5$     & 93& 97& 99& 100& 100& 92& 96& 99& 100& 100\\
              &                 & $v>0$ & $\lambda=1$       & 90& 95& 98& 99& 100& 93& 97& 99& 100& 100\\
              & \multicolumn{3}{l|}{Regularized LCM} & 82& 92& 95& 94& 94& 77& 79& 82& 84& 85\\
              & \multicolumn{3}{l|}{Unrestricted LCM}             & 0& 0& 0& 0& 0& 100& 100& 100& 100& 100\\
              & \multicolumn{3}{l|}{RLAM, Delta-matrix}           & 11& 12& 14& 15& 13& 89& 88& 87& 86& 87\\
&&&&&&&&&&&&& \\
11            & ESRLCM          & $v=0$ & $\lambda=0.5$     & 86& 91& 94& 97& 98& 91& 96& 98& 100& 100\\
              &                 & $v=0$ & $\lambda=1$       & 80& 86& 91& 94& 96& 93& 96& 99& 100& 100\\
              &                 & $v>0$ & $\lambda=0.5$     & 87& 93& 97& 99& 100& 91& 96& 98& 100& 100\\
              &                 & $v>0$ & $\lambda=1$       & 82& 90& 95& 98& 99& 93& 96& 99& 100& 100\\
              & \multicolumn{3}{l|}{Regularized LCM} & 77& 87& 90& 92& 91& 70& 71& 73& 76& 78\\
              & \multicolumn{3}{l|}{Unrestricted LCM}             & 0& 0& 0& 0& 0& 100& 100& 100& 100& 100\\
&&&&&&&&&&&&& \\
16            & ESRLCM          & $v=0$ & $\lambda=0.5$     & 74& 84& 89& 93& 95& 89& 94& 98& 100& 100\\
              &                 & $v=0$ & $\lambda=1$       & 59& 74& 83& 88& 92& 91& 95& 98& 100& 100\\
              &                 & $v>0$ & $\lambda=0.5$     & 75& 85& 92& 96& 99& 89& 95& 98& 100& 100\\
              &                 & $v>0$ & $\lambda=1$       & 51& 78& 87& 92& 96& 92& 95& 98& 100& 100\\
              & \multicolumn{3}{l|}{Regularized LCM} & 70& 85& 90& 91& 91& 66& 64& 67& 70& 74\\
              & \multicolumn{3}{l|}{Unrestricted LCM}             & 0& 0& 0& 0& 0& 100& 100& 100& 100& 100\\
              & \multicolumn{3}{l|}{RLAM, Delta-matrix}           & 3& 3& 4& 4& 4& 97& 97& 96& 97& 96\end{tabular}
}
\caption{On $200$ simulated datasets, we examine the recovery of $\theta$-restrictions. For each item, every pair of classes is considered. If $\theta_{j c}=\theta_{j c'}$ then the pair $(c,c')$ is considered `restricted' to the same value of $\theta$. Conversely when $\theta_{j c}\neq\theta_{j c'}$, this pair is considered `unrestricted'. Restriction sensitivity measures the recovery of restricted pairs: Correct Restriction Pairs / Total Restriction Pairs. Restriction specificity measures the recovery of unrestricted pairs: Unrestricted Pairs / Total Unrestricted Pairs. These measures are aggregated across items and averaged across simulations, and is based on the posterior mode choice of restrictions.}
\label{table:simsRestrictions}
\end{table}

%% file: PaperSections/0700_applications.tex
\FloatBarrier
\section{Applications}

We offer an application in education. The application is a cognitive test comprised of logic puzzles involving progressive matrices (SPM-LS) \citep{RAVEN1941}.

In the application we fit the same models as used in the simulation in Section~\ref{section:sims}. We fit each model with between $C=2$ and $C=16$ classes. We use $5$ chains . Goodness of fit is measured based on $20$-fold cross-validation (CV).

%% file: PaperSections/0720_applicationMatrices.tex
\FloatBarrier
\subsection{Progressive Matrices}

We examine a $12$-question cognitive test of logic puzzles called the Standard Progressive Matrices (SPM-LS) \citep{RAVEN1941}. Among the competing models, equivalence set RLCMs fit best with $C=6$ classes, $v=0$, and $q=0.5$ (Table~\ref{table:progMatGof}). For this best model, the class response probabilities are given in Table~\ref{table:progMatResponses}. Class $2$ is the strongest, performing well in all items. Classes $1$ and $5$ perform almost as well, but with weaknesses in items $10-12$. Class $3$ performs next best with weaknesses in items $7-12$. Class $6$ performs worst with weaknesses in all items. Class $4$ is meaningfully different from the other classes. This class struggles on some of the easier items (e.g. items $1-3$) and excels at some of the harder items (e.g. items $7,12$).

\begin{table}[]
\formattable{
\begin{tabular}{lllll}
\multicolumn{2}{l}{Model Type}        &            & Number of Classes & LogLikelihood, Cross Validated \\\hline
ESRLCM         & $v=0$                & $q=0.5$    & 6                 & -2,744.2                       \\
ESRLCM         & $\bar{v}=1.68$       & $q=0.5$    & 6                 & -2,745.8                       \\
\multicolumn{3}{l}{Unrestricted LCM}               & 5                 & -2,754.7  \\
\multicolumn{3}{l}{Regularized LCM, Frequentist} & 7                & -2,756.7                       \\
\multicolumn{3}{l}{RLAM, Delta-matrix}             & 8                 & -2,757.7                       \\
                  
\end{tabular}
}
\caption{Progressive Matrices Application. Best models by type. Twenty-fold cross validation is performed. For each model the number of classes $C \in \{2,\cdots,16\}$ is chosen based on cross validation. The ESRLCM hyperparameter $q\in \{0.5,1\}$ is chosen in the same way.}
\label{table:progMatGof}
\end{table}

\begin{table}[]
\formattable{
\begin{tabular}{lrrrrrr}
& \multicolumn{6}{c}{Class Response Probabilities ($\%$)}              \\
Item       & Class=2 & 1   & 5   & 3   & 4   & 6   \\\hline
1              & 83 & 83 & 83 & 83 & 38 & 38 \\
2              & 97 & 97 & 97 & 97 & 56 & 56 \\
3              & 94 & 94 & 88 & 88 & 13 & 27 \\
4              & 99 & 66 & 99 & 66 & 99 & 6  \\
5              & 99 & 82 & 99 & 82 & 99 & 7  \\
6              & 94 & 94 & 94 & 52 & 11 & 11 \\
               &    &    &    &    &    &    \\
7              & 90 & 90 & 71 & 24 & 90 & 24 \\
8              & 87 & 87 & 55 & 7  & 55 & 7  \\
9              & 92 & 92 & 36 & 25 & 25 & 25 \\
               &    &    &    &    &    &    \\
10             & 80 & 5  & 33 & 5  & 33 & 5  \\
11             & 82 & 6  & 18 & 6  & 6  & 18 \\
12             & 66 & 12 & 12 & 12 & 66 & 12 \\
               &    &    &    &    &    &    \\
Average:       & 89 & 67 & 65 & 46 & 49 & 20 \\
Class Size ($\pi$): & 33 & 10 & 30 & 11 & 5  & 11
\end{tabular}
}
\caption{Progressive Matrices Application. Response probabilities under ESRLCM with $C=6$ classes, $q=0.5$, and $v=0$.}
\label{table:progMatResponses}
\end{table}

%% file: PaperSections/1000_conclusion.tex
\FloatBarrier

\section{Concluding Remarks}

We have introduced a new type of regularizing LCM called the equivalence set RLCM. We have shown in simulations and applications that this model is successful in capturing common class response probabilities, and in doing so achieves a strong goodness of it.

\section{Funding}

The authors gratefully acknowledge the financial support of the National Science Foundation grants SES 1758631 and SES 21-50628.

%% file: Appendix/0200_restrictions.tex
\FloatBarrier

\section{Equivalence Set Restrictions - Identifiability}\label{appendix:esRestrictions}

In this subsection, we show generic identifiability of equivalence set RLCMs. We start with some definitions and references to past work.

\begin{definition}
The Kruskal rank, $\text{rank}_{\kappa} \mymatrix{M}$, is the largest number $k$ where every set of $k$ columns of $\mymatrix{M}$ are linearly independent.
\end{definition}

\begin{definition}
Let $\khatrirao$ represent the Khatri-Rao product, applied column-wise. 
\end{definition}

\begin{theorem}\label{thm:Kruskal}
[Kruskal] Let vector $\myvec{W}$ be: 
\begin{align}
\myvec{W} 
& = [\mymatrix{A}^{(0)} \khatrirao \mymatrix{A}^{(1)} \khatrirao \mymatrix{A}^{(2)}]\ \myvec{1}
\label{eq:triprod}
\\
& = [\mymatrix{B}^{(0)}\khatrirao \mymatrix{B}^{(1)} \khatrirao \mymatrix{B}^{(2)}]\ \myvec{1}
\nonumber
\end{align}
where $\mymatrix{A}^{(k)}$ and $\mymatrix{B}^{(k)}$ are $w_{k} \times w$ matrices. If:
\begin{align}
& \text{rank}_{\kappa} \mymatrix{A}^{(0)}
+ \text{rank}_{\kappa} \mymatrix{A}^{(1)}
+ \text{rank}_{\kappa} \mymatrix{A}^{(2)}
\geq 2 w + 2.
\end{align}
Then $[\mymatrix{A}^{(0)},\mymatrix{A}^{(1)},\mymatrix{A}^{(2)}]$ equals $[\mymatrix{B}^{(0)},\mymatrix{B}^{(1)},\mymatrix{B}^{(2)}]$ up to simultaneous permutation and rescaling of columns.
\end{theorem}

\begin{proof}
See \cite{Kruskal1977}.
\end{proof}

\begin{definition}
For items $\mathcal{J}$, let matrix $\mymatrix{\mathbbm{T}}$ be a pattern probability matrix. Matrix $\mymatrix{\mathbbm{T}}$ is a $C \times \prod_{j \in \mathcal{J}} m_{j}$ containing has one row per class and one column per response pattern to items $\mathcal{J}$. Each cell gives the conditional probability of this class producing the given response pattern.
\end{definition}

\begin{theorem}
Assume that pattern probability matrix $\mymatrix{\mathbbm{T}}$ is formed by the row-wise Kronecker product of conditionally independent items. If there exists any set of response probabilities where $\mymatrix{\mathbbm{T}}$ is full rank, then $\mymatrix{\mathbbm{T}}$ is almost surely full rank.
\end{theorem}

\begin{proof}
Proof by Allman, restated roughly here. For a $C \times K$ matrix $\mymatrix{\mathbbm{T}}$ let $m:=\text{min}(C,K)$ be the minimum dimension. Recall that a minor of matrix $\mymatrix{\mathbbm{T}}$ is the determinant of some $m\times m$ submatrix of $\mymatrix{\mathbbm{T}}$.

Matrix $\mymatrix{\mathbbm{T}}$ is nonzero if and only if there is a nonzero minor. We know there exists a specific $\mymatrix{\mathbbm{T}}_{0}$ where $\mymatrix{\mathbbm{T}}_{0}$ is full rank. Therefore there exists some minor on $\mymatrix{\mathbbm{T}}_{0}$ which is nonzero. Take the same submatrix and consider the minor on $\mymatrix{\mathbbm{T}}$. This submatrix determinant (i.e. minor) is a polynomial of the response probabilities of the corresponding items. We know that this polynomial has a single nonzero value corresponding to $\mymatrix{\mathbbm{T}}_{0}$. Therefore the polynomial is only zero on at most a measure zero space. Therefore $\mymatrix{\mathbbm{T}}$ is almost surely full rank.
\end{proof}

From this point on, we build lemmas in preparation of our main proof.

\begin{corollary}
\label{thm:distinctResponses}
Assume that pattern probability matrix $\mymatrix{\mathbbm{T}}$ is formed by the row-wise Kronecker product of conditionally independent items. Suppose there exists a choice of response probabilities every class loads onto a unique response: $P(\myvec{X}_{i}=\myvec{r}_{c}|c_{i}=c)=1$ with $\myvec{r}_{c} \neq \myvec{r}_{c'}$ when $c \neq c'$. Then $\mymatrix{\mathbbm{T}}$ is almost surely full rank.
\end{corollary}

\begin{theorem}\label{thm:fullRankProbabilities}
Let $\mathcal{J}$ be a subset of conditionally independent items with corresponding pattern probability matrix $\mymatrix{\mathbbm{T}}$. Let $\mymatrix{B}$ be the base class matrix of items $\mathcal{J}$. Let the number of possible response patterns of items $\mathcal{J}$ meet or exceed the number of classes: $\prod_{j \in \mathcal{J}_{k}} m_{j} \geq C$.

If there exists a merged base class matrix $\tilde{\mymatrix{B}}$ matching the following conditions then $\text{Rank}(\mymatrix{\mathbbm{T}})=C$ almost surely. 
\begin{enumerate}
\item For each column $\tilde{\myvec{B}}_{j}$, the number of base classes cannot exceed the number of item response levels: $\mathcal{B}_{j}\leq m_{j}$.
\item Every row $\tilde{\myvec{B}}_{c\cdot}^{(k)}$ must be unique.
\end{enumerate}
\end{theorem}

\begin{proof}
Construct $\tilde{\myvec{B}}$ as described. We will prove by constructing a set of response probabilities $\tilde{\mymatrix{\mathbbm{T}}}$ and using Corollary~\ref{thm:distinctResponses}.

By \#1 we know that the number of base classes is less than or equal to the number of response levels. Therefore for each base class and item we can let the $r$'th response level correspond with the $r$'th base class: $\tilde{\theta}_{b j,r=b} = P(X_{i j}=r|\tilde{b}_{j i}=b)= 1$. For this choice of $\tilde{\theta}$ each row $\tilde{\myvec{\mathbbm{T}}}_{c\cdot}$ directly corresponds with a single pattern: $\tilde{\mathbbm{T}}_{c r} \in \{0,1\}$. By \#2 each row $\tilde{\myvec{B}}_{c\cdot}$ is unique, and therefore each row $\tilde{\myvec{\mathbbm{T}}}_{c\cdot}$ is unique, corresponding with a different pattern. Therefore this choice of $\tilde{\mymatrix{\mathbbm{T}}}$ is full rank, and by Corollary~\ref{thm:distinctResponses} it follows that $\tilde{\mymatrix{\mathbbm{T}}}$ is almost surely full rank. Since $\tilde{\mymatrix{\mathbbm{T}}}$ is a special case of $\mymatrix{\mathbbm{T}}$, it also follows that $\mymatrix{\mathbbm{T}}$ is almost surely full rank.
\end{proof}

\begin{remark}
In Theorem~\ref{thm:fullRankProbabilities}, relabeling $\myvec{B}_{j}$ to $\tilde{\myvec{B}}_{j}$ has the effect of mapping each base class to a specific response of item $j$. By ensuring that $\tilde{\mymatrix{B}}$ has unique rows we show that each class can be mapped to a unique response pattern. This almost surely guarantees us a full rank matrix under Corollary~\ref{thm:distinctResponses}.
\end{remark}

\thmEsIdentifiability*

\begin{proof}
Following Kruskal and Allman, we know that generic identifiability holds if almost surely:
\begin{align*}
\text{rank}_{k} \mymatrix{\mathbbm{T}}_{1}+\text{rank}_{k} \mymatrix{\mathbbm{T}}_{2}+\text{rank}_{k} \mymatrix{\mathbbm{T}}_{3} \geq 2C+2
\end{align*}

By Theorem~\ref{thm:fullRankProbabilities}, we know that the first two tripartitions are full rank and therefore have Kruskal rank $C$. For the third tripartition, we know that every pair of classes have distinct base classes. Therefore every pair almost surely have distinct probabilities and have a Kruskal rank is atleast 2.
\end{proof}

\thmQidentifiability*

\begin{proof}
Any row of a Q-matrix $\myvec{Q}_{j\cdot}$ can be translated into a corresponding base class vector $\myvec{B}_{j}$. Since our responses are binary the relabeled base classes  $\tilde{\myvec{B}}_{j}$ must only have two values. For each row $\myvec{Q}_{j\cdot}$ let us create a relabeled $\tilde{\myvec{Q}}_{j\cdot}$ from which we will generate $\tilde{\myvec{B}}_{j}$. Necessarily $\tilde{\mymatrix{Q}} \leq \mymatrix{Q}$ elementwise. We let $\tilde{\mymatrix{M}}_{1}=\tilde{\mymatrix{M}}_{2}=\mymatrix{I}$. For this choice of $\tilde{\mymatrix{M}}_{k}$ every row of $\tilde{\myvec{B}}_{j}$ is unique. Additionally the number of response patterns equals the number of classes. For the third submatrix $\tilde{\mymatrix{M}}_{3}$ we know that each trait is active for atleast one item. As a result each row of original base classes $\myvec{B}_{c,\cdot}$ is unique for these items.
\end{proof}

%% file: Appendix/0300_dependentBeta.tex
\FloatBarrier
\newpage

\section{Repelled Beta Distribution}\label{appendix:repelledBeta}

\subsection{Monotone Repelled Beta Distribution}

\begin{definition}[Monotone Repelled Beta]
The monotone repelled beta density is given by:
\begin{align}
P(\myvec{\rho} \sim \text{MonRepelledBeta}(\mymatrix{\alpha}, v))
&\propto \prod_{k=1}^{M} [\rho_{k}^{\alpha_{1 k}-1}(1-\rho_{k})^{\alpha_{2 k}-1}] \prod_{k=2}^{M} (\rho_{k}-\rho_{k-1})^{v}
\label{eq:MdepThetaJoint}
\end{align}
on the allowed region where $0<\rho_{1} \leq \rho_{2}\leq \cdots \leq \rho_{m} < 1$ and with $v\geq 0$. 
\end{definition}

\begin{remark}
The monotone repelled beta is designed to encourage $\rho_{k+1}-\rho_{k} \gg 0$. In particular $P(\rho_{k+1}=\rho_{k})=0$ in density when $v>0$.
\end{remark}

\begin{definition}\label{def:delta}
By convention let $\rho_{0} := 0$. Let $\delta_{k} := \rho_{\oo{k}}-\rho_{\oo{k-1}}$ for $1 \leq k \leq M$ and $\delta_{M+1}:= 1 - \sum_{k=1}^{M} \delta_{k} = 1-\rho_{M}$. By construction $\myvec{\delta}\in\text{Simplex}_{M+1}$ where $\text{Simplex}_{k}:=\{\myvec{z}\in (0,1)^{k}:\sum_{i=1}^{k} z_{i}=1\}$. This definition implies that $\rho_{k} := \sum_{j=1}^{k} \delta_{j}$. 
\end{definition}

\begin{theorem}
\label{thm:rhoDiffs} 
Let $\myvec{\rho} \sim \text{MonRepelledBeta}(\mymatrix{\alpha}, v)$ with $\alpha_{1 k},\alpha_{2 k} \in \mathbb{N}$ for all $k$. The density of $\myvec{\delta} \in \text{Simplex}_{M+1}$ as given in Definition~\ref{def:delta} is:
\begin{align}
P(\myvec{\delta}) &\propto \prod_{k=1}^{M} [(\sum_{r=1}^{k} \delta_{r})^{\alpha_{1 k}-1} (\sum_{r=k+1}^{M+1} \delta_{r})^{\alpha_{2 k}-1}] (\prod_{k=2}^{M} \delta_{k}^{v})
\label{eq:depBetaDiff}
\end{align}
which represents a mixture of one or more Dirichlet distributions.
\end{theorem}

\begin{proof}
We begin by finding $P(\myvec{\delta})$ using change of variables. Note that there is a bijection between $\myvec{\rho}$ and $\myvec{\delta}$. We can convert between $\myvec{\rho}$ and $\myvec{\delta}$ via the following linear transformation:

\begin{align*}
\left[\begin{array}{c}
\delta_{1}   \\
\delta_{2}   \\
\delta_{3}   \\
\vdots       \\
\delta_{M}   \\
\end{array}\right]
&=
\begin{blockarray}{ccccccc}
\rho_{{1}} & \rho_{{2}} & \rho_{{3}} & \rho_{{4}} & \cdots & \rho_{{M-1}} & \rho_{{M}}       \\
\begin{block}{[ccccccc]}
    1        & 0         & 0         & 0         & \cdots & 0            & 0                \\
    -1        & 1         & 0         & 0         & \cdots & 0            & 0                \\
0         & -1        & 1         & 0         & \cdots & 0            & 0                \\
\vdots    & \vdots    & \vdots    & \vdots    & \ddots & \vdots       & \vdots     & \vdots \\
0         & 0         & 0         & 0         & \cdots & -1           & 1                \\
\end{block}
\end{blockarray}
\left[\begin{array}{c}
\rho_{{1}}   \\
\rho_{{2}}   \\
\rho_{{3}}   \\
\vdots       \\
\rho_{{M}} \\
\end{array}\right] 
 =
\mymatrix{W}
\myvec{\rho}
\\ 
\end{align*}

with $M \times M$ matrix $\mymatrix{W}$. Note that $|\mymatrix{W}|=1=|\mymatrix{W}^{-1}|$. Let $\myvec{\rho}:=\myvec{\rho}(\myvec{\delta})$ be a function of delta. The resulting change of variables is:
\begin{align*}
P(\myvec{\delta})
&= P(\myvec{\rho})
\left| \frac{d \myvec{\rho}}{d \myvec{\delta}} \right|
= P(\myvec{\rho})
\left| \mymatrix{W^{-1}} \right|
= P(\myvec{\rho})
\end{align*}
Equation \eqref{eq:depBetaDiff} can be finished by substituting $\myvec{\delta}$ into $P(\myvec{\rho})$.

A mixture of Dirichlets can be observed by expanding the terms of \eqref{eq:depBetaDiff} (i.e. using the multinomial theorem).
\end{proof}

\begin{corollary}\label{thm:delta1s} When $\alpha_{1 k} = \alpha_{2 k}=1$ for all $k$ then:
\begin{align*}
\myvec{\delta} & \sim \text{Dirichlet}( [1, v+1, \cdots, v+1, 1]^{\top}),\;\myvec{\delta}\in \text{Simplex}_{M+1}\\
E[\delta_{k}] &= \left\{\begin{array}{cl}
\frac{1}{ (M-1) (v+1) + 2 } & : k \in \{1, M+1\} \\
\frac{v+1}{ (M-1) (v+1) + 2 } & : 1 < k < M+1 
\end{array}\right.\\
P(\myvec{\rho} \sim \text{MonRepelledBeta}(\mymatrix{1}, v))
&= \frac{\Gamma((M-1)(v+1)+2)}{\Gamma^{M-1}(v+1)} \prod_{k=2}^{M} (\rho_{k}-\rho_{k-1})^{v}\\
E[\rho_{k}] &= \frac{1 + (v+1)(k-1)}{ (M-1) (v+1) + 2 }
\end{align*}
\end{corollary}

\begin{definition}\label{def:repBetaNormalizer}
Let $d(\mymatrix{\alpha},v)$ be the normalizing constant for $\text{MonRepelledBeta}(\mymatrix{\alpha}, v)$ such that:
\begin{align}
P(\myvec{\rho} \sim \text{MonRepelledBeta}(\mymatrix{\alpha}, v))
&= d(\mymatrix{\alpha},v) \prod_{k=1}^{M} [\rho_{k}^{\alpha_{1 k}-1}(1-\rho_{k})^{\alpha_{2 k}-1}] \prod_{k=2}^{M} (\rho_{k}-\rho_{k-1})^{v}
\end{align}
\end{definition}

\subsection{Repelled Beta Distribution}

\begin{definition}
Let $\myvec{\pi}$ denote a permutation of $M$ elements: $\{\pi_{k}:k\}=\mathbb{Z}_{M}$. Let $\myvec{A}_{\myvec{\pi}}:=[A_{\pi_{1}},\cdots,A_{\pi_{M}}]^{\top}$ denote the permuted elements of a vector and $\mymatrix{B}_{\myvec{\pi}}:=[\myvec{B}_{\pi_{1}},\cdots,\myvec{B}_{\pi_{M}}]$ denote the permuted columns of a matrix.
\end{definition}

\begin{definition}[Repelled Beta]\label{def:repbeta2}
Let $\myvec{\rho} \sim \text{RepelledBeta}(\mymatrix{\alpha}, v) \in (0,1)^{M}$ follow the repelled beta distribution. The repelled beta density is given a mixture of permuted monotone repelled betas:
\begin{align}
P(\myvec{\rho} = \myvec{q})
&\propto \sum_{\myvec{\pi}:\{\pi_{k}:k\}=\mathbb{Z}_{M}} d(\mymatrix{\alpha}_{\myvec{\pi}},v)^{-1}
P(\text{MonRepelledBeta}(\mymatrix{\alpha}_{\myvec{\pi}},v) = \myvec{q}_{\myvec{\pi}})
\end{align}
with $d(\mymatrix{\alpha},v)$ as given in Definition~\ref{def:repBetaNormalizer}.
\end{definition}

\begin{remark}
Note that the mixed components in Definition~\ref{def:repbeta2} are almost surely mutually exclusive due to monotonicity. For nearly any choice of $\myvec{q}$ there is only one permuted monotone repelled beta which could generate it.
\end{remark}

\begin{theorem}
The density of $\myvec{\rho} \sim \text{RepelledBeta}(\mymatrix{\alpha}, v)$ with $2\times M$ matrix $\mymatrix{\alpha}$ is given by:
\begin{align}
P(\myvec{\rho} = \myvec{q})
&\propto \prod_{k=1}^{M} [q_{k}^{\alpha_{1 k}-1}(1-q_{k})^{\alpha_{2 k}-1}] \prod_{k=2}^{M} (q_{\oo{k}}-q_{\oo{k-1}})^{v}
\end{align}
with subscript $\oo{k}$ denoting the $k$'th order statistic.
\end{theorem}

\begin{proof} Let $c \in \mathbb{R}$ be the normalizing constant for this repelled beta. Assume every component of $\myvec{q}$ is unique; this removes only a measure zero space.
\begin{align*}
P(\myvec{\rho} = \myvec{q})
&= c \sum_{\myvec{\pi}:\{\pi_{k}:k\}=\mathbb{Z}_{M}} d(\mymatrix{\alpha}_{\myvec{\pi}},v)^{-1}
P(\text{MonRepelledBeta}(\mymatrix{\alpha}_{\myvec{\pi}},v) = \myvec{q}_{\myvec{\pi}}) \\
&= c \sum_{\myvec{\pi}:\{\pi_{k}:k\}=\mathbb{Z}_{M}} \prod_{k=1}^{M} [q_{\pi_{k}}^{\alpha_{1 \pi_{k}}-1}(1-q_{\pi_{k}})^{\alpha_{2 \pi_{k}}-1}] \prod_{k=2}^{M} (q_{\pi_{k}}-q_{\pi_{k}-1})^{v} I(q_{\pi_{k}}<\cdots<q_{\pi_{k}})\\
&= c \sum_{\myvec{\pi}:\{\pi_{k}:k\}=\mathbb{Z}_{M}} \prod_{k=1}^{M} [q_{k}^{\alpha_{1 k}-1}(1-q_{k})^{\alpha_{2 k}-1}] \prod_{k=2}^{M} (q_{\oo{k}}-q_{\oo{k-1}})^{v} I(q_{\pi_{k}}<\cdots<q_{\pi_{k}})\\
&= c \prod_{k=1}^{M} [q_{k}^{\alpha_{1 k}-1}(1-q_{k})^{\alpha_{2 k}-1}] \prod_{k=2}^{M} (q_{\oo{k}}-q_{\oo{k-1}})^{v}
\end{align*}    
\end{proof}

\begin{remark}
No order on $\myvec{\rho}$ is imposed on the repelled beta distribution. When $v=0$, the repelled beta distribution is equivalent to the distribution of independent betas. As $v$ grows, the elements of $\myvec{\rho}$ are increasingly dependent and tend further apart.
\end{remark}

\thmdepThetaPrior*

\begin{proof}
Let $c$ be the normalizing constant of this repelled beta distribution. Note that every column of $\mymatrix{\alpha}$ is identical. From Definition~\ref{def:repbeta2}, it follows that:
\begin{align*}
\int P(\myvec{\rho} = \myvec{q}) d\myvec{q}
&= c \sum_{\myvec{\pi}:\{\pi_{k}:k\}=\mathbb{Z}_{M}} d(\mymatrix{\alpha}_{\myvec{\pi}},v)^{-1}
\int P(\text{MonRepelledBeta}(\mymatrix{\alpha}_{\myvec{\pi}},v) = \myvec{q}_{\myvec{\pi}}) d\myvec{q}\\
&= c \sum_{\myvec{\pi}:\{\pi_{k}:k\}=\mathbb{Z}_{M}} d(\mymatrix{\alpha},v)^{-1}
(1) \\
&= \frac{c\ M!}{d(\mymatrix{\alpha},v)} = 1
\end{align*}
By Corollary~\ref{thm:delta1s} we have $d(\mymatrix{\alpha},v)=\frac{\Gamma((M-1)(v+1)+2)}{\Gamma^{M-1}(v+1)}$. Substituting this and solving for $c$ finishes the proof.
\end{proof}

\subsection{Other Properties}

\thmdepThetaConjugacyb*

\begin{proof} Proof is analogous to the conjugacy of beta and Bernouli distributions.
\begin{align*}
P(\mymatrix{Z}|\myvec{\rho})
&=
\prod_{k} \prod_{i}^{l_{k}}\rho_{k}^{I(Z_{i k}=1)}(1-\rho_{k})^{I(Z_{i k}=0)}
\\
& =
\prod_{k}\rho_{k}^{\sum_{i}^{l_{k}} I(Z_{i k}=1)}(1-\rho_{k})^{\sum_{i}^{l_{k}} I(Z_{i k}=0)} \\
& =\prod_{k}\rho_{k}^{n_{k 1}}(1-\rho_{k})^{n_{k 2}} 
\\
P(\myvec{\rho}|\mymatrix{Z})
& \propto P(\myvec{\rho})
P(\mymatrix{Z}|\myvec{\rho})
\\
&\propto \prod_{k}^{M} [\rho_{k}^{ n_{k 1}+\alpha_{k 1}-1}(1-\rho_{k})^{n_{k 2}+\alpha_{k 2}-1}] \prod_{k=2}^{M} (\rho_{\oo{k}}-\rho_{\oo{k-1}})^{v}
\end{align*}

\end{proof}

\begin{remark}
Conjugacy between the monotone repelled beta and Bernouli responses is essentially identical.
\end{remark}

%% file: Appendix/0500_posteriors.tex
\FloatBarrier
\newpage

\section{Posterior Approximation}\label{appendix:posteriors}

\subsection{StickBreaking}\label{appendix:posteriorsStickbreaking}

In this subsection we examine the distribution of the number of base classes $\nbaseclass_{j}$ collapsed on the stick-breaking probability $\myvec{\eta}_{j}$. When collapsed in this way $\nbaseclass_{j}$ becomes a multinoulli distribution with probabilities depending on $\mymatrix{\alpha}_{j}^{(\eta)}$.

\begin{remark}\label{remark:beta}
The Beta function has the following properties:
\begin{align*}
\beta(\myvec{v}) & := \frac{\prod_{k} \Gamma(v_{k})}{\Gamma(\sum_{k} v_{k})} \\
\int_{\text{Simplex}_{M}} \prod_{b=1}^{M} x^{\alpha_{b}-1} d\myvec{x} & = \beta(\myvec{\alpha})
\end{align*}
Furthermore let $\myvec{w} = [0,\cdots,0,1,0,\cdots,0]^{\top}$ with a value of one in the $b'$ component. Then we have the following proportion:
\begin{align}
\frac{\beta(\myvec{v}+\myvec{w})}{\beta(\myvec{v})}
& 
= \frac{\Gamma(\sum v_{k})}{\Gamma(1+\sum v_{k})} \frac{\Gamma(v_{b'}+1) \prod_{k\neq b'} \Gamma(v_{k})}{\prod_{k} \Gamma(v_{k})}
= \frac{v_{b'}}{\sum v_{k}}
\label{eq:betaRatio}
\end{align}

\end{remark}

\begin{theorem}\label{thm:nbasecollapsed}When collapsed on $\myvec{\eta}_{j}$, the number of base classes is distributed:
\begin{align*}
P(\nbaseclass_{j}=m) 
&=\left( \prod_{b}^{m-1} \frac{\alpha^{(\eta)}_{1 b j}}{\alpha^{(\eta)}_{1 b j}+\alpha^{(\eta)}_{2 b j}} \right) \left(1-\frac{\alpha^{(\eta)}_{1 m j}}{\alpha^{(\eta)}_{1 m j}+\alpha^{(\eta)}_{2 m j}}\right)^{I(m<C)} \\
&=\left( \prod_{b}^{m-1} E[\eta_{b j}] \right) (1-E[\eta_{m j}])^{I(m<C)}
\end{align*}
\end{theorem}

\begin{proof} We omit subscript $j$ below for notational convenience.
\begin{align*}
P(\nbaseclass=m)
&= \int P(\myvec{\eta})P(\nbaseclass_{j}=m|\myvec{\eta}) d\myvec{\eta} \\
&= \int \prod_{b}^{C-1} \left(\frac{1}{\beta(\myvec{\alpha}^{(\eta)}_{b})} \eta_{b}^{\alpha^{(\eta)}_{1 b}-1} (1-\eta_{b})^{\alpha^{(\eta)}_{2 b}-1}\right)\left[(\prod_{k}^{m-1} \eta_{k}) (1-\eta_{m})^{I(m<C)}\right] d\myvec{\eta} \\
&= 
\prod_{b}^{C-1} \left(\frac{1}{\beta(\myvec{\alpha}^{(\eta)}_{b})}\right) \int 
\prod_{b}^{m-1} \left(\eta_{b}^{\alpha^{(\eta)}_{1 b}+1-1} (1-\eta_{b})^{\alpha^{(\eta)}_{2 b}-1}\right) 
\\&\quad\quad \cdot
\left(\eta_{m}^{\alpha^{(\eta)}_{1 m}-1} (1-\eta_{m})^{\alpha^{(\eta)}_{2 m}+I(m<C)-1}\right)
\prod_{m+1}^{C-1} \left(\eta_{b}^{\alpha^{(\eta)}_{1 b}-1} (1-\eta_{b})^{\alpha^{(\eta)}_{2 b}-1}\right)d\myvec{\eta} \\
&= 
\prod_{b}^{C-1} \left(\frac{1}{\beta(\myvec{\alpha}^{(\eta)}_{b})}\right) 
\left( \prod_{b}^{m-1} \beta(\alpha^{(\eta)}_{1 b}+1,\alpha^{(\eta)}_{2 b})\right) 
\\&\quad\quad \cdot
\beta(\alpha^{(\eta)}_{1 m},\alpha^{(\eta)}_{2 m}+I(m<C))
\left(\prod_{m+1}^{C-1} \beta(\alpha^{(\eta)}_{1 b},\alpha^{(\eta)}_{2 b})\right) \\
&= 
\left( \prod_{b}^{m-1} \frac{\alpha^{(\eta)}_{1 b}}{\alpha^{(\eta)}_{1 b}+\alpha^{(\eta)}_{2 b}} \right) \left(\frac{\alpha^{(\eta)}_{2 m}}{\alpha^{(\eta)}_{1 m}+\alpha^{(\eta)}_{2 m}}\right)^{I(m<C)}
\end{align*}
The last line above uses equation \eqref{eq:betaRatio}.
\end{proof}

\begin{corollary}\label{eq:existsProbNbase}
For any $\myvec{\zeta} \in \text{Simplex}_{C}$ there exists $\myvec{\alpha}_{1 j}^{(\eta)},\cdots,\myvec{\alpha}_{C j}^{(\eta)}$ such that $\nbaseclass_j\sim\text{Categorical}(\myvec{\zeta})$ with $P(\nbaseclass_{j}=m)=\zeta_{m}$ for all $m$.
\end{corollary}

\begin{proof} Proof amounts roughly to the interchangeability of StickBreaking and Categorical distributions. Let $\hat{\eta}_{b j}:= E[\eta_{b j}] = \frac{\alpha_{1 b j}}{\alpha_{1 b j}+\alpha_{2 b j}}$. From Theorem~\ref{thm:nbasecollapsed}, we see that $P(\nbaseclass_{j}=m)$ follows a StickBreaking distribution with parameters $\hat{\eta}_{b j}$. The values $\myvec{\zeta}$ can be viewed as parameters for a Categorical distribution. StickBreaking and Categorical distributions are different parameterizations of the same model. Therefore for any choice of $\myvec{\zeta}$ there exists an equivalent set of $\hat{\myvec{\eta}}_{j}$. For any choice of $\hat{\myvec{\eta}}_{j}$ there are many choices of $\mymatrix{\alpha}_{j}^{(\eta)}$.
\end{proof}

\begin{corollary}
For any $\myvec{\lambda} \in \mathbb{R}_{+}^{C}$, there exists an $\mymatrix{\alpha}_{j}^{(\eta)}$ such that the prior of $\myvec{B}_{j}$ collapsed on $\myvec{\eta}_{j}$ is given by:
\begin{align}
P(\myvec{B}_{j}) = \frac{\lambda_{\nbaseclass_{j}}}{\sum_{k}^{C} \stirling{C}{k}  \lambda_{k}} \propto \lambda_{\nbaseclass_{j}} \label{eq:basevecProptoProb2}
\end{align}
\end{corollary}

\begin{proof}
Follows immediately from Corollary~\ref{eq:existsProbNbase} with $\zeta_{k} = (\stirling{C}{k}  \lambda_{k})/(\sum_{k}^{C} \stirling{C}{k}  \lambda_{k})$.
\end{proof}

\subsection{MCMC Update of Base Classes}

\subsubsection{When $v=0$ fixed}\label{appendix:mcmcmBaseClassV0}

When $v=0$, we can update base classes $\myvec{B}_{j}$ with a Gibbs step collapsed on $\myvec{\theta}'_{j}$. In our Gibbs update we choose a random component of $\myvec{B}_{j}$ and update that base class. The Gibbs update works in the usual fashion using the collapsed response likelihoods given by:

\begin{theorem} Collapsed on $\theta_{j}'$, we have:
\begin{align*}
P(\myvec{x}_{j}|\myvec{B}_{j},v=0,\myvec{c}) 
&= \beta(\alpha_{1},\alpha_{2})^{-\nbaseclass_{j}} \prod_{b=1}^{\nbaseclass_{j}} \beta(\alpha_{1}+n^{(\theta)}_{j b 1},\alpha_{2}+n^{(\theta)}_{j b 2})
\\
n^{(\theta)}_{j b r} & := \sum_{i} I(B_{c_{i} j}=b,x_{i j}=1-r)
\end{align*}
\end{theorem}

\begin{proof}
Follows standard proof for collapsing a response probability on a categorical response. Subscript $j$ omitted for simplicity. Let $b_{i} := B_{c_{i} j}$ be the base class of observation $i$.
\begin{align*}
P(\myvec{x}|\myvec{B},v=0,\myvec{c}) 
& = \int P(\myvec{\theta}'|v=0) \prod_{i} P(\myvec{x}_{j}|c_{i},\myvec{\theta}',\myvec{B}_{j}) d\myvec{\theta}'\\
& = \prod_{b=1}^{\nbaseclass} \int P(\theta'_{b} \sim \text{Beta}(\alpha_{1},\alpha_{2})) \prod_{i} \theta_{b}^{\prime I(b_{i}=b)I(x_{i}=1)}(1-\theta_{b}')^{I(b_{i}=b)I(x_{i}=0)} d\theta'_{b}\\
& = \beta(\alpha_{1},\alpha_{2})^{-\nbaseclass} \prod_{b=1}^{\nbaseclass} \int \theta_{b}^{\prime \alpha_{1}+\sum_{i} I(b_{i}=b)I(x_{i}=1)}(1-\theta_{b}')^{\alpha_{2}+\sum_{i} I(b_{i}=b)I(x_{i}=0)} d\theta'_{b}\\
&= \beta(\alpha_{1},\alpha_{2})^{-\nbaseclass} \prod_{b=1}^{\nbaseclass_{j}} \beta(\alpha_{1}+n^{(\theta)}_{b 1},\alpha_{2}+n^{(\theta)}_{b 2}) 
\end{align*}
\end{proof}

\subsubsection{When $v>0$}

We do a reversible jump \citep{GREEN1995} update of $\myvec{B}_{j}$.

\subheader{Proposal Distribution}: Our reversible jump starts in state $(\myvec{B}_{j},\myvec{\theta}'_{j})$ and we propose state $(\myvec{\tilde{B}}_{j},\myvec{\tilde{\theta}}'_{j})$. We use the following random process to generate proposed $(\myvec{\tilde{B}}_{j},\myvec{\tilde{\theta}}'_{j})$. The decision of accepting the proposal is deferred to the next paragraph. First we initialize $(\myvec{\tilde{B}}_{j},\myvec{\tilde{\theta}}'_{j})=(\myvec{B}_{j},\myvec{\theta}'_{j})$. Next we update $\myvec{\tilde{B}}_{j}$. This update follows exactly the process described in Section~\ref{appendix:mcmcmBaseClassV0} as though we were running a Gibbs update with $v=0$. Unlike the previous section, this simply builds a proposal, which will not necessarily be accepted. This update changes a single component of $\myvec{\tilde{B}}_{j}$ denoted $\tilde{B}_{\tilde{c} j}$ at index $\tilde{c}$. Next we update update two components of $\theta'_{b j}$ corresponding to the initial ($b = B_{\tilde{c} j}$) and new base classes ($b = \tilde{B}_{\tilde{c} j}$) of the modified component. These proposed $\tilde{\theta}'$ are generated from independent draws from the beta distribution:
\begin{align}
\text{Proposal}(\tilde{\theta}'_{b j}|\tilde{B}_{j})
& \sim \text{Beta}(\alpha_{1}+\tilde{n}_{j b 1},\alpha_{2}+\tilde{n}_{j b 2}) \label{eq:revJumpThetaProp}\\
\tilde{n}^{(\theta)}_{j b r} & := \sum_{i} I(\tilde{B}_{c_{i} j}=b,x_{i j}=1-r) \nonumber
\end{align}

\subheader{Acceptance Probability} If the number of base classes does not change, this reduces to a standard metropolis step. Below we consider the reversible jump step when the number of base classes increases (the decreasing case is symmetric). An increase in base classes occurs when the initial base class contains multiple classes ($|E_{j b}|>1$ for $b= B_{\tilde{c} j}$), and component $\tilde{c}$ moves to an entirely new base class ($|\tilde{E}_{j b}|=1$ for $b= \tilde{B}_{\tilde{c} j}$). Without loss of generality assume the initial number of base classes be $m$, and the modified component moves from $B_{\tilde{c} j}=m$ to $\tilde{B}_{\tilde{c} j}=m+1$. Let $U_{1},U_{2}$ be the proposed values of $\tilde{\theta}'_{b j}$ for $b \in \{m+1,m\}$ respectively. Then our proposal is 
\begin{align*}
g(\myvec{\theta}'_{j}, U_{1}, U_{2}) &= (\myvec{\tilde{\theta}}'_{1:M-1,j}=\myvec{\theta}'_{1:M-1,j}, \tilde{\theta}'_{m}=U_{1}, \tilde{\theta}'_{m+1}=U_{2}, U_{3}=\theta'_{m j})    
\end{align*}
The reverse operation is given by:
\begin{align*}
g^{-1}(\myvec{\tilde{\theta}}'_{j}, U_{3}) 
&= (\myvec{\theta}'_{1:M-1,j}=\myvec{\tilde{\theta}}'_{1:M-1,j}, \theta'_{m}=U_{3}, U_{1}=\tilde{\theta}'_{m}, U_{2}=\tilde{\theta}'_{m+1})
\end{align*}
Our proposal functions $g(\cdot)$ operate by relabeling elements. Note that we have \quote{dimension matching} between $(\myvec{\theta}'_{j}, U_{1}, U_{2})$ and $(\myvec{\tilde{\theta}}'_{j}, U_{3})$. Additionally the determinant of the Jacobian is $|\frac{d g(\myvec{\theta},\myvec{U})}{d(\myvec{\theta},\myvec{U})}| = 1$. Using the usual formula for reversible jump, our acceptance probability is:
\begin{align*}
\alpha 
& = \min\left\{1, \frac{P(\myvec{\tilde{B}}_{j},\myvec{\tilde{\theta}}'_{j}|\myvec{x}_{j},v,\myvec{c})P(U_{3})}{P(\myvec{B}_{j},\myvec{\theta}'_{j}|\myvec{x}_{j},v,\myvec{c})P(U_{1},U_{2})} \left|\frac{d g(\myvec{\theta}_{j}',\myvec{U})}{d(\myvec{\theta}_{j}',\myvec{U})}\right|\right\} \\
& = \min\left\{1, \frac{P(\myvec{\tilde{B}}_{j})P(\myvec{\tilde{\theta}}'_{j}|\myvec{\tilde{B}}_{j})P(\myvec{x}_{j}|v,\myvec{c},\myvec{\tilde{\theta}}'_{j},\myvec{\tilde{B}}_{j})P(U_{3})}{P(\myvec{B}_{j})P(\myvec{\theta}'_{j}|\myvec{B}_{j})P(\myvec{x}_{j}|v,\myvec{c},\myvec{\theta}'_{j},\myvec{B}_{j})P(U_{1})P(U_{2})}\right\} \\
\end{align*}

\subsection{MCMC Update of v}

\begin{definition}[Triangular Distribution] The triangular distribution $\text{Triangular}(a,b,c)$ has a sample space of $[a,c]$. The density of the triangular distribution [$(x,f(x))$] is formed by linearly interpolating $(a,0)$, $(b,\frac{2}{b-1})$, and $(c,0)$. In this way the density of the triangular distribution forms a triangle on $(a,c)$ with mode value at $b$.
\end{definition}

We use a metropolis step to update $v$. Our metropolis proposal uses a triangular distribution $\text{Triangular}(a=0,b=\hat{b},c=\text{MaxV})$ with $\hat{b}$ as the current/existing value of $v$. The full conditional of $v$ is given by:
\begin{align*}
P(v|\mymatrix{\theta}')
&\propto P(v)P(\mymatrix{\theta}'|v,\mymatrix{B}) \\
& \propto \left[v^{d_{1}} e^{d_{2} v} I(0 < v < \text{MaxV})\right] \left[\prod_{j} \frac{\Gamma((\nbaseclass_{j}-1)(v+1)+2)}{\nbaseclass_{j}!\Gamma^{\nbaseclass_{j}-1}(v+1)} \prod_{k=2}^{\nbaseclass_{j}} (\theta_{\oo{k} j}'-\theta_{\oo{k-1} j}')^{v}\right] \\
&= \left(\prod_{j} \frac{\Gamma((\nbaseclass_{j}-1)(v+1)+2)}{\nbaseclass_{j}!\Gamma^{\nbaseclass_{j}-1}(v+1)}\right) v^{d_{1}} e^{(d_{2}-D_{2}) v} I(0 < v < \text{MaxV}) \numberthis \label{eq:vPosterior} \\
D_{2} & := - \sum_{j=1,k=2}^{j=J,k=\nbaseclass_{j}} \ln (\theta_{\oo{k} j}'-\theta_{\oo{k-1} j}') \geq 0
\end{align*}

Recall that $d_{1},d_{2}>0$ to promote regularization.

\subsection{Other Updates}

The updates to class membership $\myvec{c}$ and class prior $\myvec{\pi}$ follow the usual LCM conjugacies (e.g. see \citet{Li2018}).

For $\myvec{\theta}'_{j}$, we have a conjugate prior (see Theorem~\ref{thm:thmdepThetaConjugacyb}).

%% file: Appendix/0600_simulations.tex
\FloatBarrier
\newpage

\section{Simulation Studies}\label{appendix:simulations}

\begin{table}[]
\formattable{
\begin{tabular}{l|rrrrr}
Base Classes & Class=1 & 2 & 3 & 4 & 5 \\\hline
Item1        & 0       & 1 & 2 & 2 & 0 \\
Item2        & 2       & 3 & 1 & 0 & 3 \\
Item3        & 0       & 1 & 1 & 0 & 1 \\
Item4        & 0       & 1 & 1 & 0 & 1 \\
Item5        & 2       & 2 & 1 & 0 & 1 \\
Item6        & 0       & 1 & 1 & 0 & 0 \\
Item7        & 3       & 1 & 0 & 2 & 1 \\
Item8        & 2       & 0 & 3 & 1 & 1 \\
Item9        & 2       & 3 & 0 & 1 & 3 \\
Item10       & 0       & 0 & 1 & 2 & 0 \\
Item11       & 0       & 2 & 1 & 0 & 2 \\
Item12       & 1       & 0 & 2 & 3 & 1 \\
Item13       & 0       & 0 & 1 & 1 & 1 \\
Item14       & 0       & 1 & 2 & 1 & 2 \\
Item15       & 0       & 1 & 0 & 1 & 1 \\
Item16       & 0       & 0 & 1 & 1 & 0 \\
Item17       & 2       & 0 & 1 & 1 & 1 \\
Item18       & 0       & 1 & 0 & 1 & 0 \\
Item19       & 1       & 2 & 0 & 3 & 2 \\
Item20       & 0       & 1 & 1 & 0 & 0 \\
Item21       & 1       & 2 & 2 & 0 & 0 \\
Item22       & 2       & 0 & 3 & 1 & 1 \\
Item23       & 1       & 0 & 2 & 1 & 2 \\
Item24       & 1       & 0 & 1 & 0 & 1 \\
Item25       & 1       & 1 & 0 & 0 & 0 \\
Item26       & 2       & 1 & 0 & 1 & 1 \\
Item27       & 1       & 1 & 0 & 0 & 0 \\
Item28       & 0       & 2 & 1 & 3 & 3 \\
Item29       & 2       & 1 & 2 & 0 & 2 \\
Item30       & 0       & 3 & 1 & 2 & 0 \\
Item31       & 2       & 1 & 1 & 0 & 2 \\
Item32       & 1       & 3 & 2 & 0 & 2
\end{tabular}
}
\caption{Simulations: Base classes used in data generation when $C\in \{4,5\}$. When $C<5$, the first $C$ columns are used. Response probabilities are evenly spaced between $1/(2 \nbaseclass_{j})$ when $B_{c j}=0$ and $1-1/(2 \nbaseclass_{j})$ when $B_{c j}=\nbaseclass_{j}-1$. Each class has equally likely prior probability.}
\end{table}

\begin{table}[]
\formattable{
\begin{tabular}{l|rrrrrrrrrrrrrrrr}
Base Classes & \multicolumn{16}{c}{Class} \\
Item & 1 & 2 & 3 & 4 & 5 & 6 & 7 & 8 & 9 & 10 & 11 & 12 & 13 & 14 & 15 & 16 \\\hline
1        & 2       & 3 & 0 & 2 & 1 & 4 & 4 & 1 & 2 & 0  & 2  & 4  & 0  & 1  & 3  & 4  \\
2        & 1       & 2 & 4 & 0 & 0 & 2 & 3 & 1 & 4 & 4  & 0  & 1  & 2  & 2  & 3  & 0  \\
3        & 7       & 1 & 0 & 5 & 3 & 4 & 2 & 6 & 2 & 7  & 5  & 1  & 0  & 3  & 6  & 4  \\
4        & 0       & 1 & 1 & 0 & 0 & 0 & 1 & 1 & 0 & 1  & 0  & 1  & 1  & 0  & 0  & 1  \\
5        & 2       & 0 & 2 & 1 & 0 & 1 & 1 & 2 & 1 & 0  & 0  & 2  & 0  & 2  & 1  & 1  \\
6        & 1       & 1 & 2 & 0 & 0 & 0 & 1 & 2 & 0 & 1  & 1  & 2  & 2  & 1  & 2  & 0  \\
7        & 2       & 4 & 2 & 0 & 4 & 1 & 1 & 3 & 1 & 3  & 4  & 0  & 1  & 2  & 0  & 3  \\
8        & 0       & 2 & 6 & 4 & 5 & 1 & 7 & 3 & 2 & 4  & 6  & 0  & 5  & 3  & 7  & 1  \\
9        & 0       & 2 & 2 & 1 & 0 & 4 & 4 & 3 & 1 & 3  & 2  & 0  & 4  & 1  & 0  & 2  \\
10       & 2       & 5 & 4 & 5 & 2 & 3 & 1 & 0 & 5 & 2  & 4  & 1  & 5  & 0  & 3  & 0  \\
11       & 1       & 1 & 0 & 0 & 0 & 1 & 0 & 1 & 0 & 1  & 1  & 1  & 0  & 0  & 1  & 0  \\
12       & 5       & 1 & 3 & 0 & 4 & 6 & 7 & 2 & 3 & 5  & 7  & 4  & 1  & 0  & 6  & 2  \\
13       & 2       & 1 & 3 & 1 & 2 & 0 & 3 & 0 & 1 & 2  & 2  & 3  & 3  & 0  & 1  & 0  \\
14       & 1       & 3 & 3 & 2 & 0 & 4 & 4 & 2 & 0 & 4  & 3  & 1  & 2  & 3  & 2  & 1  \\
15       & 2       & 5 & 6 & 4 & 0 & 1 & 3 & 0 & 0 & 1  & 6  & 5  & 3  & 5  & 2  & 4  \\
16       & 1       & 2 & 0 & 0 & 1 & 2 & 0 & 1 & 1 & 0  & 0  & 2  & 0  & 2  & 2  & 1  \\
17       & 0       & 2 & 1 & 0 & 2 & 1 & 1 & 0 & 1 & 1  & 2  & 0  & 0  & 1  & 2  & 2  \\
18       & 1       & 1 & 0 & 0 & 1 & 0 & 1 & 0 & 0 & 1  & 1  & 0  & 1  & 0  & 0  & 1  \\
19       & 0       & 3 & 1 & 3 & 5 & 0 & 4 & 2 & 0 & 3  & 5  & 1  & 4  & 4  & 3  & 2  \\
20       & 5       & 0 & 3 & 6 & 5 & 2 & 1 & 4 & 1 & 3  & 0  & 5  & 6  & 4  & 2  & 4  \\
21       & 2       & 1 & 0 & 0 & 1 & 2 & 0 & 1 & 2 & 0  & 1  & 0  & 1  & 2  & 0  & 2  \\
22       & 0       & 1 & 3 & 5 & 2 & 3 & 4 & 4 & 0 & 5  & 3  & 1  & 2  & 0  & 4  & 5  \\
23       & 6       & 4 & 2 & 3 & 0 & 5 & 1 & 1 & 3 & 6  & 5  & 0  & 1  & 5  & 2  & 4  \\
24       & 2       & 0 & 3 & 0 & 1 & 2 & 1 & 3 & 1 & 3  & 0  & 0  & 2  & 2  & 1  & 3  \\
25       & 5       & 6 & 0 & 1 & 2 & 7 & 4 & 3 & 3 & 5  & 0  & 1  & 4  & 2  & 7  & 6  \\
26       & 1       & 1 & 1 & 0 & 0 & 1 & 0 & 0 & 1 & 0  & 0  & 0  & 0  & 1  & 1  & 1  \\
27       & 2       & 0 & 5 & 3 & 6 & 4 & 3 & 1 & 3 & 4  & 5  & 6  & 0  & 4  & 2  & 1  \\
28       & 1       & 1 & 0 & 1 & 0 & 1 & 0 & 0 & 0 & 0  & 0  & 0  & 1  & 1  & 1  & 1  \\
29       & 1       & 0 & 3 & 2 & 3 & 2 & 1 & 0 & 3 & 2  & 0  & 3  & 1  & 0  & 1  & 2  \\
30       & 3       & 1 & 2 & 2 & 0 & 1 & 0 & 3 & 2 & 2  & 0  & 3  & 0  & 1  & 1  & 3  \\
31       & 0       & 1 & 1 & 0 & 2 & 4 & 5 & 3 & 0 & 4  & 1  & 2  & 4  & 5  & 3  & 0  \\
32       & 2       & 3 & 1 & 0 & 3 & 1 & 0 & 2 & 1 & 0  & 2  & 3  & 3  & 0  & 1  & 2 
\end{tabular}
}
\caption{Simulations: Base classes used in data generation when $C \in \{8,11,16\}$. When $C<16$, the first $C$ columns are used. Response probabilities are evenly spaced between $1/(2 \nbaseclass_{j})$ when $B_{c j}=0$ and $1-1/(2 \nbaseclass_{j})$ when $B_{c j}=\nbaseclass_{j}-1$. Each class has equally likely prior probability.}
\end{table}

%% file: Appendix/0710_applicationFractionSubtraction.tex
\FloatBarrier
\subsection{Application: Fraction Subtraction}


We examine a $20$-question math test given to $536$ middle school students \citep{Tatsuoka1984AnalysisOE}. This test examines a student's competence in subtracting two fractions. This data is available in the R CDM package \citep{jsscdm}. 

Among the competing models, an equivalence set RLCM fit best with $C=9$ classes, $v=0$, and $q=1$ (Table~\ref{table:fractionGof}). For the model with nine classes, five classes represent $90\%$ of the population. The remaining four `nuisance' classes together represent only $10\%$ of the population. 

The class response probabilities are shown in Table~\ref{table:fractionResponses}. Inspecting this reveals four basic types of items: Whole number minus fraction (e.g. $3 - 2 1/5$), smaller minus larger fraction (e.g. $3 \frac{1}{2} - 2 \frac{3}{2}$), common denominators (e.g. $\frac{5}{3} - \frac{3}{4}$), and subtracting numerators (e.g. $\frac{6}{7} - \frac{4}{7}$). In increasing levels of skill the five main classes are $4$, $5$, $2$, $9$, $3$. The four nuisance classes tend to have mixed performance among problems of the same type. For instance nuisance class $8$ does well in `smaller minus larger fraction' problems except for item $13$. Item $13$ is the hardest problem in that category. For full details see Table~\ref{table:fractionResponses}.

\begin{table}[]
\formattable{
\begin{tabular}{lllll}
Model         &                     &        & \# of Classes & LogLikelihood, Cross Validated \\\hline
ESRLCM            & $v=0$                & $q=1$       & 9        & -4,362.8          \\
ESRLCM            & $\bar{v}=1.67$       & $q=1$       & 8        & -4,363.8          \\
\multicolumn{3}{l}{RLAM, Delta-matrix}             & 16       & -4,372.3          \\
\multicolumn{3}{l}{Unrestricted LCM}               & 8        & -4,378.3          \\
\multicolumn{3}{l}{Regularized LCM,   Frequentist} & 7        & -4,455.3        
\end{tabular}
}
\caption{Fraction Subtraction Application. Best models by type. Twenty-fold cross validation is performed. For each model the number of classes $C \in \{2,\cdots,16\}$ is chosen based on cross validation. For the RLAM, $C=32$ was considered as well. The ESRLCM hyperparameter $q\in \{0.5,1\}$ is chosen in the same way.}
\label{table:fractionGof}
\end{table}

\begin{table}[]
\formattable{
\begin{tabular}{rlrrrrr:rrrr}
\multicolumn{1}{l}{} &                   & \multicolumn{9}{c}{Class Response   Probabilities (Percent)}              \\
\multicolumn{2}{l}{Item}                 & class=3 & 9    & 2    & 5    & 4    & 8    & 1    & 6    & 7  \\\hline
\multicolumn{2}{l}{Whole Number Minus Fraction} &      &      &      &      &      &      &      \\
7:                   & 3 - 2 1/5         & 89    & 30 & 41 & 2  & 2  & 89 & 2  & 89 & 30  \\
15:                  & 2 - 1/3           & 93    & 49 & 49 & 5  & 5  & 93 & 5  & 93 & 5   \\
19:                  & 4 - 1 4/3         & 84    & 13 & 13 & 1  & 1  & 84 & 1  & 1  & 1   \\
 &                   &         &      &      &      &      &      &      &      \\
 \multicolumn{2}{l}{Smaller Minus Larger Fraction} &      &      &      &      &      &      &      \\
4:                   & 3 1/2 - 2 3/2     & 89    & 89 & 27 & 10 & 27 & 89 & 52 & 52 & 52   \\
10:                  & 4 4/12 - 2 7/12   & 79    & 79 & 12 & 2  & 2  & 79 & 38 & 2  & 12  \\
11:                  & 4 1/3 - 2 4/3     & 93    & 93 & 8  & 8  & 8  & 93 & 93 & 8  & 8   \\
13:                  & 3 3/8 - 2 5/6     & 66    & 66 & 12 & 1  & 1  & 12 & 1  & 1  & 1   \\
17:                  & 7 3/5 - 4/5       & 94    & 71 & 4  & 4  & 4  & 71 & 71 & 71 & 4   \\
18:                  & 4 1/ 10 - 2 8/ 10 & 84    & 84 & 39 & 10 & 2  & 84 & 84 & 39 & 2   \\
20:                  & 4 1/3 - 1 5/3     & 89    & 65 & 2  & 2  & 2  & 89 & 65 & 2  & 2   \\
 &                   &         &      &      &      &      &      &      &      \\
\multicolumn{2}{l}{Common Denominators} &      &      &      &      &      &      &      \\
1:                   & 5/3 - 3/4         & 88    & 88 & 88 & 8  & 4  & 8  & 4  & 4  & 88  \\
2:                   & 3/4 - 3/8         & 96    & 96 & 96 & 6  & 6  & 6  & 6  & 6  & 89  \\
3:                   & 5/6 - 1/9         & 89    & 89 & 89 & 2  & 2  & 2  & 2  & 2  & 64  \\
5:                   & 4 3/5 - 3 4/10    & 89    & 66 & 66 & 39 & 25 & 39 & 25 & 39 & 66  \\
 &                   &         &      &      &      &      &      &      &      \\
\multicolumn{2}{l}{Subtracting Numerators} &      &      &      &      &      &      &      \\
6:                   & 6/7 - 4/7         & 97    & 97 & 97 & 86 & 15 & 86 & 86 & 86 & 97  \\
8:                   & 2/3 - 2/3         & 95    & 95 & 72 & 54 & 54 & 95 & 54 & 54 & 72  \\
9:                   & 3 7/8 - 2         & 86    & 64 & 64 & 76 & 32 & 86 & 18 & 32 & 32  \\
12:                  & 1 1/8 - 1/8       & 94    & 94 & 81 & 81 & 13 & 94 & 94 & 94 & 13  \\
14:                  & 3 4/5 - 3 2/5     & 94    & 94 & 94 & 79 & 5  & 94 & 79 & 94 & 5   \\
16:                  & 4 5/7 - 1 4/7     & 93    & 93 & 79 & 79 & 6  & 79 & 79 & 65 & 6   \\
 &                   &         &      &      &      &      &      &      &      \\
\multicolumn{2}{l}{Average:} & 89 & 76 & 52 & 28 & 11 & 69 & 43 & 42 & 32 \\
\multicolumn{2}{l}{Class Size ($\pi$):}           & 30    & 11 & 13 & 17 & 18 & 2  & 3  & 1  & 3 
\end{tabular}
}
\caption{Fraction Subtraction Application. Response probabilities under ESRLCM model with $C=9$ classes, $q=1$, and $v=0$.}
\label{table:fractionResponses}
\end{table}